\documentclass[letterpaper,journal]{IEEEtran}
\usepackage{amsmath,amsfonts}
\usepackage{algorithmic}
\usepackage{algorithm}
\usepackage{array}
\usepackage[caption=false,font=normalsize,labelfont=sf,textfont=sf]{subfig}
\usepackage{textcomp}
\usepackage{stfloats}
\usepackage{url}
\usepackage{verbatim}
\usepackage{graphicx}
\usepackage{cite}

\usepackage{booktabs}
\usepackage{amsmath}
\usepackage{amssymb}
\usepackage{balance}
\usepackage{hyperref}
\usepackage{multirow}
\usepackage{bbding}
\usepackage{orcidlink}

\usepackage{tikz}
\usepackage{amsmath}
\newcommand{\sqdiamond}[1][fill=black]{\tikz [x=1.2ex,y=1.85ex,line
	width=.1ex,line join=round, yshift=-0.285ex] \draw  [#1]  (0,.5) -- (.5,1) --
	(1,.5) -- (.5,0) -- (0,.5) -- cycle;}%
\newcommand{\mydiamond}[1][fill=black]{\mathop{\raisebox{-0.205ex}{$\sqdiamond[#1]$}}}

\begin{document}
\title{Regularized Schr\"odinger Bridge via Distortion-Perception Perturbation for High-Fidelity Speech Enhancement}

\author{Qing Yao\,\orcidlink{0009-0001-2172-8004}, Lijian Gao, Qirong Mao\,\orcidlink{0000-0002-0616-4431}, and Ming Dong \,\orcidlink{0000-0001-8133-7809}

\thanks{This work has been accepted for publication in the
\textit{IEEE/ACM Transactions on Audio, Speech, and Language Processing}. 
This work was supported in part by the National Natural Science Foundation of China (Grant Nos. 62576155, 62506144, and 62176106), the Natural Science Foundation of Jiangsu Province (Grant No. SH2025108), and the Postgraduate Research \& Practice Innovation Program of Jiangsu Province (Grant No. KYCX25\_4234). (The corresponding authors: Qirong Mao; Ming Dong)

Qing Yao, Lijian Gao and Qirong Mao are with the School of Computer Science and Communication Engineering, Jiangsu University, Zhenjiang, Jiangsu 212013, China, also with the Jiangsu Engineering Research Center of Big Data Ubiquitous Perception and Intelligent Agriculture Applications, Zhenjiang, Jiangsu 212013, China, and also with the Provincial Key Laboratory of Computational Intelligence and New Technologies in Low-Altitude Digital Agriculture, Zhenjiang, Jiangsu 212013, China. (email: $qyao@stmail.ujs.edu.cn$; $\{ljgao, mao\_qr\}@ujs.edu.cn$)

Ming Dong is with the Department of Computer Science, Wayne State University, Detroit, MI 48202, USA. (email: $mdong@wayne.edu$)

Digital Object Identifier: \href{https://doi.org/10.1109/TASLPRO.2026.3717234}{10.1109/TASLPRO.2026.3717234}.
}}

\markboth{IEEE/ACM Transactions on Audio, Speech, and Language Processing}%
{Yao \MakeLowercase{\textit{et al.}}:
Regularized Schrödinger Bridge via Distortion--Perception Perturbation
for High-Fidelity Speech Enhancement}

\IEEEpubid{%
  \begin{minipage}{\textwidth}
  \fontsize{7}{8}\selectfont
  \centering
    \textcopyright~2026 IEEE. Personal use of this material is permitted.  Permission from IEEE must be obtained for all other uses, in any current or future media, including reprinting/republishing this material for advertising or promotional purposes, creating new collective works, for resale or redistribution to servers or lists, or reuse of any copyrighted component of this work in other works.
  \end{minipage}%
}

\maketitle

\begin{abstract}
Speech enhancement (SE) requires high-fidelity reconstruction of clean speech that preserves linguistic and paralinguistic cues while maintaining high perceptual quality. Recently, Schr\"odinger Bridge (SB), a family of diffusion-based generative models, has advanced SE by bridging degraded and clean speech distributions in a principled formulation, enabling higher-quality reconstructions with fewer sampling steps. However, diffusion-based SE methods still face two challenges: (1) the fidelity-realism tradeoff, where they often prioritize perceptual realism encouraged by the learned speech prior, at the expense of fidelity; and (2) the exposure bias issue, where iterative multi-step sampling causes early-step prediction errors to accumulate along the sampling trajectory and degrade enhanced speech quality. In this paper, we analyze standard SB training and show that it induces a systematic prediction drift, which biases the multi-step trajectory and amplifies error accumulation. To address this, we propose Regularized Schr\"odinger Bridge (RSB) for high-fidelity SE, a generative approach that reconciles fidelity and realism while mitigating exposure bias. RSB regularizes training with a Distortion-Perception Perturbation that constructs time-varying targets by interpolating between clean speech and posterior-mean estimates, and trains the network on perturbed intermediate states to correct toward the ground truth progressively. By simulating inference-time prediction errors, this perturbation mitigates the training-inference mismatch and thereby alleviates exposure bias. It also injects posterior-mean estimates as fidelity-preserving guidance, thereby improving reconstruction fidelity. Experiments on the WSJ0 corpus and the VoiceBank+DEMAND dataset demonstrate that RSB improves reconstruction fidelity over strong diffusion-based generative methods, yielding a favorable fidelity-realism tradeoff and reducing exposure bias.
\end{abstract}


\section{Introduction}
\IEEEPARstart{S}{peech} enhancement~(SE) is a crucial inverse problem in audio processing that seeks to recover clean speech from degraded signals corrupted by noise or reverberation. Fundamentally, perceptual realism and reconstruction fidelity are two essential criteria for evaluating SE approaches: realism concerns the perceived naturalness of the enhanced speech, whereas fidelity reflects reconstruction accuracy. The enhanced speech should sound natural while faithfully preserving fine-grained linguistic content and paralinguistic attributes (e.g., timbre or pitch). These requirements make SE particularly challenging because it is inherently ill-posed, where multiple valid reconstructions correspond to the same measurement. The main challenge is to select the desired estimate from these candidate reconstructions. Consequently, existing approaches can be broadly categorized into \emph{predictive} and \emph{generative methods}~\cite{lemercier2023storm}, each representing a distinct family of plausible solutions in the solution space.

\emph{Predictive} methods~\cite{wang2022tfgridnet, fu2021metricganplus, chao2024semamba, lu2023mpsenet} typically learn a deterministic mapping from degraded to target signals via end-to-end supervised training~\cite{ongie2020dlt, whang2021dsr}. While such methods are effective in suppressing background noise and achieving high speech fidelity, they often average multiple plausible reconstructions, yielding overly smoothed outputs that lose fine-grained details~\cite{lemercier2023adg, lemercier2023storm}. In contrast, \emph{generative} methods aim to implicitly or explicitly model the underlying distribution of clean speech to produce natural and perceptually plausible estimates on the data manifold, but they may lead to higher sample-level reconstruction errors and introduce artifacts such as vocalizations and breathing~\cite{lemercier2023adg}. Recent advancements in diffusion-based generative modeling~\cite{ho2020ddpm, song2021sgm, zhou2023ddbm, liu2023i2sb,wang2024siibb} provide SE with a strong generative prior and enable high-quality reconstructions through iterative stochastic sampling, thereby substantially elevating the state of the art in generative SE~\cite{lu2021diffuse, lu2022cdiffuse, welker2022sgm,richter2023sgmseplus}. Despite encouraging advances, diffusion-based SE methods still struggle to (1) balance perceptual realism and reconstruction fidelity, and (2) mitigate exposure bias in multi-step sampling.

\IEEEpubidadjcol
First, diffusion-based SE methods often prioritize distributional realism for perceptual quality, which can come at the expense of reconstruction fidelity. This phenomenon is commonly formalized as the \emph{distortion-perception (DP) tradeoff}~\cite{blau2017pdt, freirich2021dpt, adrai2023dot, zhou2025cdt, ohayon2025pmrf} in ill-posed inverse problems such as SE. This tradeoff captures the fundamental tension between two competing objectives, namely \emph{perception} and \emph{distortion}. Here, \emph{perception} measures the distributional discrepancy between enhanced and clean speech, which reflects perceptual realism. In contrast, \emph{distortion} denotes point-wise reconstruction error relative to the clean speech rather than perceived speech distortion, and is typically quantified by mean squared error (MSE), which directly reflects speech fidelity. Theoretically, since these two objectives cannot be simultaneously optimized~\cite{freirich2021dpt, khan2025fmse}, the optimal tradeoffs lie on the Pareto frontier of the DP plane (referred to as the DP-optimal frontier), consisting of solutions that minimize distortion at a given level of perceptual quality. In practice, diffusion-based SE methods can improve perceptual realism at the cost of reconstruction fidelity, which increases point-wise distortion and yields solutions that deviate from the DP-optimal frontier. Recently, hybrid SE approaches~\cite{qiu2023srtnet, lemercier2023storm, zhang2025pguse} integrate predictive and generative models into a pipeline, in which a diffusion model refines the predictive estimate. While such designs improve this tradeoff in practice, unreliable predictive estimates can substantially limit the final reconstruction quality.

Second, the iterative sampling procedure of these methods at inference time inevitably introduces the \emph{exposure bias} problem~\cite{ning2023ipr, li2023aeb, ning2023eeb, ren2024msds}. This issue arises from a systematic mismatch between training and inference. During training, the network is optimized to learn the diffusion dynamics while being conditioned on ground-truth intermediate states. During inference, however, it is conditioned on its own previously generated samples instead. As a result, even small prediction errors in the early steps can propagate along the sampling chain and gradually accumulate, causing the actual sampling trajectory to deviate markedly from the true trajectory~\cite{ning2023ipr}. One feasible approach to mitigate this issue is to expose the model to simulated prediction errors during training~\cite{ning2023ipr}, thereby reducing the mismatch between training and inference. By doing so, the network gradually learns to recover from imperfect inputs with accumulated prediction errors during inference, leading to improved robustness.

Recently, Schr\"odinger bridge (SB)~\cite{chen2022ltsb, liu2023i2sb}, an extension of score-based diffusion models, has emerged as a promising generative SE framework. By adopting optimal transport theory, SB learns the transport between clean and degraded speech distributions, improving both quality and sampling efficiency~\cite{jukic2024sbse}. However, relative to predictive methods that optimize point-wise error, SB still exhibits a noticeable gap in reconstruction fidelity, as reflected by signal-fidelity metrics such as SI-SDR~\cite{leroux2019sdr}. In addition, SB remains susceptible to exposure bias, as substantial error accumulation is observed in our experiments.

In this paper, we conduct a thorough examination of the standard SB training objective~\cite{jukic2024sbse} in the context of SE. Notably, to learn the reverse bridge dynamics, a common approach is to train a data-prediction network (referred to as the SB model throughout this paper) that directly estimates the clean signal from the intermediate states. Since multi-step sampling inevitably incurs prediction errors, we analyze the inference-time behavior of the SB model via these prediction errors and reveal that the standard objective induces a systematic prediction drift along the sampling trajectory. Empirically, this drift moves samples away from the DP-optimal frontier, improving perceptual quality but increasing distortion. Moreover, because inference iteratively uses the current model prediction to predict the next state, these errors can feed back into subsequent inputs, enlarging the training-inference mismatch. Such drift is not accounted for by the standard SB training. Therefore, there is a need for a principled regularization that both optimizes the DP balance and mitigates exposure bias during iterative sampling.

To address the aforementioned limitations, we propose the \emph{Regularized Schr\"odinger Bridge~(RSB)}, a generative SE method tailored for high-fidelity SE, which integrates a regularized training strategy that balances perceptual quality and reconstruction fidelity in enhanced speech. Overall, RSB regularizes SB training with our proposed DP Perturbation. It encourages the SB model to produce predictions that are highly faithful to the degraded speech at early sampling steps, and then gradually transitions toward a natural realization of clean speech at later steps, ultimately yielding estimates with both strong signal fidelity and high perceptual realism. At the same time, this regularized training exposes the model to trajectory deviations during training, mitigating the training-inference mismatch and thereby reducing exposure bias.

Specifically, unlike prior perturbation-based training methods~\cite{ning2023ipr, ren2024msds}, which inject additional Gaussian noise into training inputs to improve robustness, our DP Perturbation jointly perturbs both the training inputs and supervision targets to regularize the learned bridge transport. We start from the original SB training targets (i.e., the clean speech) and then reformulate them through a time-varying interpolation between two principled endpoints: (i) the clean speech as the perception-optimal reference, and (ii) an approximate posterior-mean estimate as the distortion-optimal reference. By injecting this posterior-mean estimate into the targets, RSB provides explicit fidelity-preserving guidance that steers the reverse dynamics toward low-distortion solutions while staying on the data manifold, thereby improving reconstruction fidelity without sacrificing realism. Importantly, we apply the same interpolation to construct perturbed training inputs that simulate inference-time states with prediction errors, reducing training-inference mismatch and improving robustness to error propagation along the sampling trajectory.

Overall, the main contributions of this work are as follows:
\begin{itemize}
    \item We study the prediction behavior of SB models through inference-time prediction errors. We find a systematic prediction drift along the sampling trajectory, which connects the DP tradeoff with error accumulation from training-inference mismatch and provides the motivation for our regularized training design.
    \item We propose a novel high-fidelity generative SE framework, RSB, which regularizes SB training via DP Perturbation. This perturbation jointly perturbs training targets and inputs through a time-varying interpolation between the clean speech and a posterior-mean estimate, thereby steering the sampling trajectory toward high-fidelity yet perceptually realistic solutions while alleviating exposure bias by reducing training-inference mismatch.
    \item We validate the efficacy of RSB across challenging SE tasks, including denoising and dereverberation. Results show that RSB improves reconstruction fidelity over the state-of-the-art diffusion-based baselines, while maintaining strong perceptual quality and mitigating exposure bias in multi-step sampling.
\end{itemize}

The rest of the paper is organized as follows. The related work is summarized in Section \ref{sec:related_work}. Section \ref{sec:method} details our proposed approach. Section \ref{sec:experiment_setup} describes the experimental setup, and Section \ref{sec:experiment_results} shows the evaluation results. Finally, the conclusion is drawn in Section \ref{sec:conclusion}.

\section{Related Work}\label{sec:related_work}
In this section, we review prior work most relevant to our study and position our contribution with respect to the DP tradeoff, exposure bias, and diffusion-based SE methods.

\subsection{Distortion-Perception Tradeoff}
In ill-posed inverse problems, the DP tradeoff~\cite{blau2017pdt} is a critical phenomenon that has captured considerable attention and is fundamental for evaluating methods that solve these problems. Specifically, these methods must trade off between \emph{distortion} and \emph{perception}, where \emph{distortion} measures the average reconstruction error of point estimates (e.g., by MSE) and thus reflects \emph{fidelity} to the target data, while \emph{perception} quantifies the divergence between the learned and true data distributions (e.g., by the Wasserstein-2 distance) and therefore reflects the perceptual realism of estimates. In general, lower distortion corresponds to reduced sampling stochasticity and higher fidelity, whereas improved perception corresponds to higher realism. The DP-optimal frontier consists of Pareto-optimal solutions that achieve the minimum distortion among all reconstructions satisfying a fixed constraint of perceptual realism. Inspired by the theoretical analysis of~\cite{blau2017pdt}, in imaging inverse problems such as image super-resolution and image deblurring, one common practice~\cite{blau2017pdt, freirich2021dpt, adrai2023dot, zhou2025cdt, ohayon2025pmrf} to approach this optimal solution involves first obtaining the minimum-distortion estimate (typically the minimum MSE estimate) and then transporting it to the target data distribution.

\subsection{Exposure Bias Problem}
Exposure bias~\cite{ning2023ipr} is a notable limitation of diffusion models that stems from discrepancies between training (conditioned on ground-truth data) and inference (conditioned on previously generated samples). As a result, even slight prediction errors can accumulate over dozens or hundreds of sampling steps, potentially causing the generated samples to move away from the true data manifold~\cite{ning2023ipr}. In denoising diffusion models, several training methods, such as input perturbation~\cite{ning2023ipr} and scheduled sampling~\cite{ren2024msds}, explicitly simulate inference-time prediction errors, thereby encouraging the network to recover from its own mistakes. In addition, sampling methods such as output scaling~\cite{ning2023eeb} and time-step shifting~\cite{li2023aeb} attempt to mitigate exposure bias without retraining by correcting accumulated prediction errors during the sampling process. However, these techniques are tailored to Gaussian noise-to-data diffusion formulations and are not directly applicable to SB, which models a data-to-data transport process.

\subsection{Diffusion-based Speech Enhancement}
Among diffusion-based SE methods, CDiffuSE~\cite{lu2022cdiffuse} is a time-domain conditional diffusion model that starts from Gaussian noise and guides the reverse process with the noisy mixture to reconstruct the clean waveform. SGMSE+~\cite{richter2023sgmseplus} proposes a score-based diffusion model in the complex spectrogram domain, which introduces a Stochastic Differential Equation~(SDE) to describe the diffusion process that transports from noisy to clean speech. SGMSE+ can generate high-quality reconstructions using only 30 steps~\cite{richter2023sgmseplus}, but it still suffers from high distortion and generative artifact issues. To reduce potential artifacts, the hybrid SE method StoRM~\cite{lemercier2023storm} proposes to cascade a predictive model with a diffusion model. This initial estimate from a predictive model is fed to the diffusion model as a guide, effectively reducing the computational burden and improving sampling quality. Recently, Jukic et al.~\cite{jukic2024sbse} introduced a tractable SB formulation for SE, delivering improved perceptual realism while maintaining a moderate computational cost. Recent studies~\cite{shang2025sbgan,lei2025fnsesbgan} incorporate Generative Adversarial Networks (GANs) into the SB framework as auxiliary perceptual supervision, using adversarial training to sharpen realistic spectral details. However, the DP tradeoff and exposure bias of SB remain open issues.

\begin{figure*}[!t]
\centering
\includegraphics[width=1.8\columnwidth]{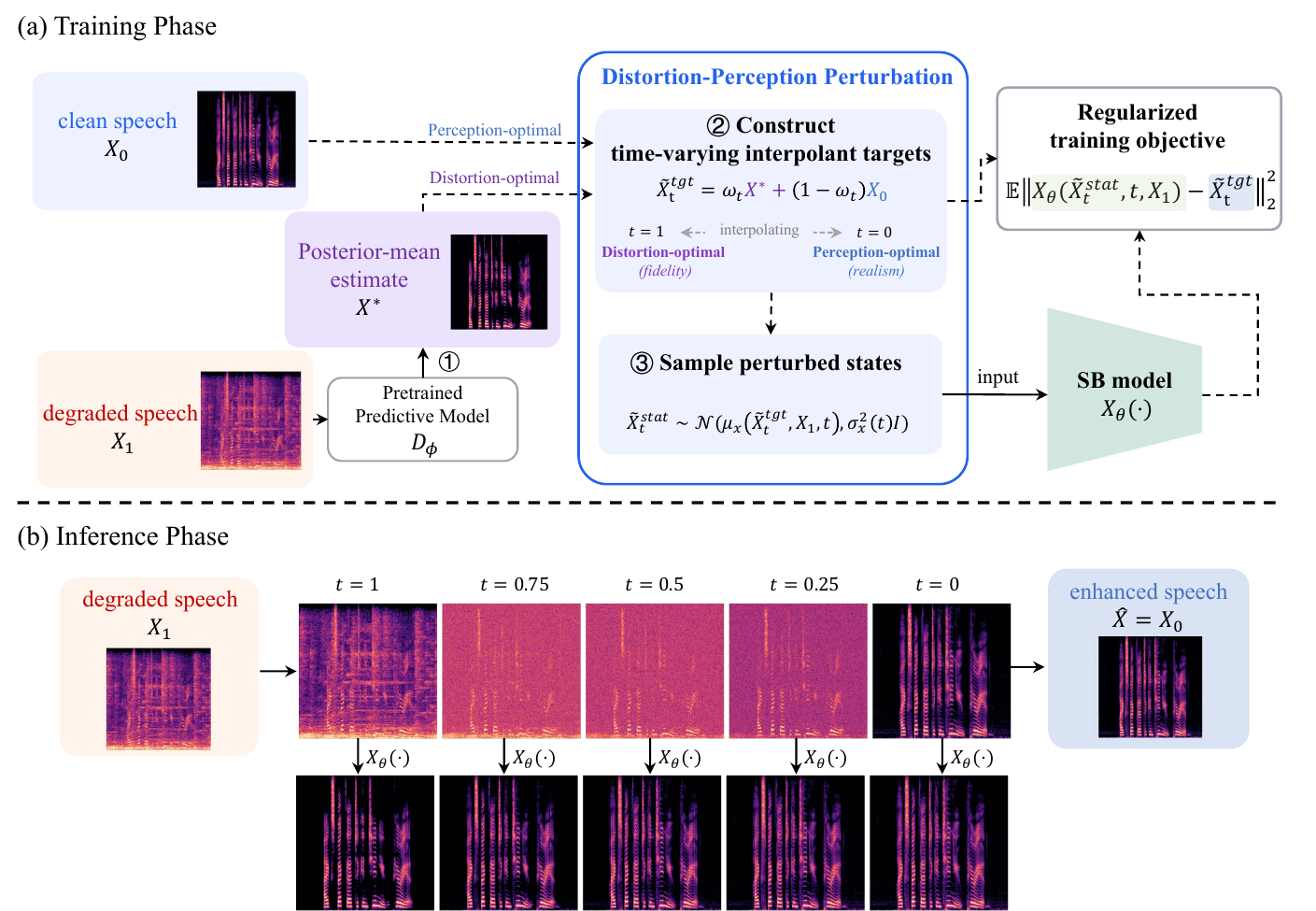}
\vspace{-3mm}
\caption{Overview of Regularized Schr\"odinger Bridge (RSB) for high-fidelity speech enhancement. (a) Training phase: Given degraded speech $X_1$ and clean speech $X_0$, a distortion-optimal reference $X^*$ (i.e., the posterior-mean estimate) is first obtained using a pretrained predictive model $D_\phi$. RSB then constructs time-varying interpolant targets $\tilde{X}_t^{\mathrm{tgt}}$ by interpolating between the distortion-optimal reference $X^*$ and the clean speech $X_0$, and samples the corresponding perturbed intermediate states $\tilde{X}_t^{\mathrm{state}}$. The SB model $X_\theta(\cdot)$ is trained to predict the perturbed targets from the perturbed states via a regularized objective, encouraging fidelity-preserving predictions at early stages and perceptually realistic predictions at later stages. (b) Inference phase: At inference time, RSB performs standard finite-step sampling, illustrating a smooth transition from low-distortion to perceptually realistic outputs.}
\vspace{-3mm}
\label{fig:rsb_overview}
\end{figure*}

\section{Regularized Schr\"odinger Bridge for High-Fidelity Speech Enhancement}\label{sec:method}
This section presents RSB for high-fidelity SE, which is illustrated in Fig.~\ref{fig:rsb_overview}. We first detail the specific formulation of SB in the context of SE, and then examine the behavior of the SB model to further analyze the DP tradeoff and exposure bias issues in SB. Finally, we present DP Perturbation in detail.

\subsection{Preliminaries: Schr\"odinger Bridge for Speech Enhancement}
In this paper, we formulate SE as an inverse problem of recovering clean speech from measurements. The clean speech $x$ is a realization of a random vector $X$ with prior distribution $p_X$. The measurement $y$ (i.e., noisy or reverberant speech signal) is a realization of a random vector $Y$. Given an observed measurement $y$, the goal of SE is to draw an estimate $\hat{x}$ from the posterior distribution $p_{\hat{X}\mid Y}(\cdot\mid y)$ such that $\hat{x}$ faithfully reconstructs the ground-truth $x$. 

\subsubsection{Formulation of Schr\"odinger Bridge}
SB~\cite{kantorovich2006otm} can be formulated as an entropy-regularized optimal transport problem between two distributions. In this paper, we formulate SB as a principled generative SE framework to find the optimal transport between the clean speech distribution $p_X$ and the measurement distribution $p_Y$. To align SB with diffusion-based generative models~\cite{chen2023sbtts}, SB can be equivalently defined by a pair of forward-backward SDEs.

We assume access to training data $\mathcal{D}=\{(x,y)\}$ drawn from the joint distribution $p_{X,Y}$, where $(x,y)$ denotes a training pair in the dataset. In the SB formulation, we treat each training pair $(x,y)$ as a coupled boundary condition. Let $X_t$ be the intermediate state of the SB process at $t\in[0,1]$ with endpoints $X_0 = x$ and $X_1 = y$, and SB admits the pair of SDEs as follows:
\begin{equation}
\label{eq:sb_sde}
    \begin{split}
        dX_t &= [\mathbf{f} + g^2(t)\nabla{\log{\Psi_t}}] dt + g(t) d\mathbf{W},\\
        dX_t &= [\mathbf{f} - g^2(t)\nabla{\log{\hat{\Psi}_t}}] dt + g(t) d\overline{\mathbf{W}},
    \end{split}
\end{equation}
where $\mathbf{f}$ and $g$ denote the drift and diffusion coefficients, respectively. $\mathbf{W}$ and $\overline{\mathbf{W}}$ are Wiener processes, and $\log{\Psi_t}$ and $\log{\hat{\Psi}_t}$ are the time-varying potentials whose gradients relate to the optimal transport path.

Generally, solving SB is intractable. However, closed-form solutions exist under the assumption of Gaussian boundary conditions~\cite{chen2023sbtts, jukic2024sbse}. Under this assumption, for the given pair of endpoints $(X_0, X_1)$, $\Psi_t$ and $\hat{\Psi}_t$ admit analytical forms:
\begin{equation}
\label{eq:sb_solution}
    \begin{split}
    \hat{\Psi}_t =\mathcal{N}(\alpha_t X_0, \alpha_t^2\sigma_t^2\mathbf{I}),
    \Psi_t =\mathcal{N}(\bar{\alpha}_t X_1, \alpha_t^2\bar{\sigma}_t^2\mathbf{I}),
    \end{split}
\end{equation}
where $\alpha_t=e^{\int_0^tf(\tau)d\tau}$, $\sigma_t^2=\int_0^t\frac{g^2(\tau)}{\alpha_\tau^2}d\tau$, $\bar{\alpha}_t=\frac{\alpha_t}{\alpha_1}$, and $\bar{\sigma}_t^2=\sigma_1^2-\sigma_t^2$ are determined by the choice of $\mathbf{f}$ and $g$. Notably, two noise schedules are derived in the tractable SB formulation, termed Variance Preserving (VP) and Variance Exploding (VE). These schedules are distinct from the common ones in standard SDEs~\cite{ho2020ddpm,song2021sgm}.

Therefore, the marginal distribution of the intermediate state $X_t$ follows a Gaussian form, expressed as
\begin{equation}
\label{eq:sb_marginal}
    p_t(X_t | X_0, X_1) = \mathcal{N}(\mu_x(X_0, X_1, t), \sigma_x^2(t) \mathbf{I}),
\end{equation}
where we have the mean $\mu_x(X_0, X_1, t) = (\alpha_t \bar{\sigma}_t^2 X_0 + \bar{\alpha}_t \sigma_t^2 X_1) /\sigma_1^2$ and the variance $\sigma_x^2(t) = {\alpha_t^2 \bar{\sigma}_t^2\sigma_t^2}/{\sigma_1^2}$.

\subsubsection{Training Objective}
Following the tractable SB, models are trained to approximate the score $\nabla{\log{\hat{\Psi}_t}}$ to learn the reverse-time SB dynamics. This score is algebraically related to the clean signal $X_0$ through the potential $\hat{\Psi}_t$ in Eq.~(\ref{eq:sb_solution}). Therefore, the widely adopted training objective is the data prediction loss~\cite{jukic2024sbse} (also called ``x-prediction''), which aims to optimize the SB model with network parameters $\theta$ to predict $X_0$ given intermediate state $X_t$ at any $t$ as,
\begin{equation}
\label{eq:sb_xpred_loss}
   \min_\theta \mathbb{E}_{X_t \sim p_t, t}[\| X_\theta(X_t, t, X_1) - X_0 \|^2_2],
\end{equation}
where $X_1$ serves as a condition of the network.

\subsubsection{Inference and Sampling}
During inference, our objective is to recover the clean speech $\hat{x}$ by generating samples from the reverse-time backward SDE defined in Eq.~(\ref{eq:sb_sde}). Specifically, the continuous-time interval $[0, 1]$ is discretized into $N$ steps, denoted by $t_0, t_1, \ldots, t_N$, where $t_0 = 1$ and $t_N = 0$. The sampling process proceeds iteratively from $t_0$ to $t_N$. Given the previously sampled state $\tilde{X}_{t_k}$ at step $t_k$ (with $k=0,1,\ldots,N-1$), the next state $\tilde{X}_{t_{k+1}}$ at $t_{k+1}<t_k$ can be obtained by the derived SDE sampler~\cite{jukic2024sbse} as,
\begin{equation}
\label{eq:sb_sde_sampler}
\begin{aligned}
\tilde{X}_{t_{k+1}} ={} & \frac{\alpha_{t_{k+1}}\sigma^2_{t_{k+1}}}{\alpha_{t_k}\sigma^2_{t_k}}\, \tilde{X}_{t_k} \\
& + \alpha_{t_{k+1}} \left(1-\frac{\sigma^2_{t_{k+1}}}{\sigma^2_{t_k}}\right) X_\theta(\tilde{X}_{t_k}, t_k, X_1) \\
& + \alpha_{t_{k+1}}\, \sigma_{t_{k+1}}\, \sqrt{1-\frac{\sigma^2_{t_{k+1}}}{\sigma^2_{t_k}}}\,\mathbf{z}
\end{aligned}
\end{equation}
where $\mathbf{z} \sim \mathcal{N}(0,\mathbf{I})$. Alternatively, the equivalent Ordinary Differential Equation (ODE) formulation of Eq.~(\ref{eq:sb_sde}) allows for sampling using an ODE sampler~\cite{jukic2024sbse}.

After iterative sampling, the final estimated clean speech $\hat{X}$, which represents the solution to SE, is defined as the state reached at the terminal time $t_N=0$. Crucially, at the terminal time $t_N=0$, we always have $\alpha_{t_N}=1$ and $\sigma_{t_N} = 0$. Consequently, both the first and the third terms in Eq.~(\ref{eq:sb_sde_sampler}) are eliminated. Then, the final sampled state $\tilde{X}_{t_N}$ is precisely the model prediction at the last sampling step, i.e.,
\begin{equation}
\hat{X} = \tilde{X}_{t_N} = X_\theta(\tilde{X}_{t_{N-1}}, t_{N-1}, X_1).
\end{equation}

\subsection{Limitations of Schr\"odinger Bridge in Speech Enhancement}
Based on the aforementioned definition of SB, we further analyze the DP tradeoff and the exposure bias of SB in SE, and then motivate our DP Perturbation.

\begin{figure}[t]
\centering
\vspace{-3mm}
\includegraphics[width = 0.75 \columnwidth]{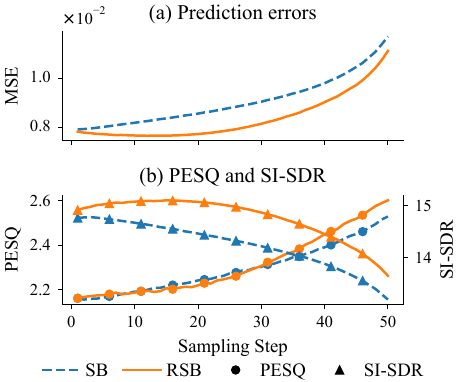}
\vspace{-3mm}
\caption{Prediction errors (a) and evaluation metrics (b) during 50-step sampling on the WSJ0+WHAM dataset. The sampling process is uniformly discretized from $1$ to $0$ over 50 steps. (a): The prediction errors at each timestep are computed using the mean squared error (MSE), $\mathbb{E}\bigl[\lVert X_{\theta}(\cdot) - X_0\rVert_2^{2}\bigr]$, between the model prediction and the ground truth. (b): PESQ measures perceived speech quality, while SI-SDR measures point-wise distortion.}
\label{fig:pred_error_trajectory}
\end{figure}

\subsubsection{Theoretical and Empirical Analysis of Prediction Errors} We begin by theoretically examining the behavior of the SB model during inference, particularly in the context of the data-prediction training objective in Eq.~(\ref{eq:sb_xpred_loss}). The SB model $X_\theta$ predicts the clean signal $X_0$ from an intermediate state $X_t$ (conditioned on $X_1$), and this prediction is algebraically related to the reverse-time score $\nabla \log \hat{\Psi}_t$ through Eq.~(\ref{eq:sb_solution}). Therefore, any discrepancy between $X_\theta$ and the ground-truth target directly perturbs the reverse-time score and, consequently, the drift term of the reverse SDE, steering the sampling direction during multi-step inference.

Specifically, at $t = 1$, the training objective reduces to a purely discriminative regression loss that, given $X_1$, minimizes the MSE between the model prediction $X_\theta(\cdot)$ and the clean target $X_0$. Consequently, during sampling, due to inevitable prediction errors, the inference-time prediction $X_\theta(X_t, X_1, t)$ around $t = 1$ is often biased toward the minimum-MSE solution, namely the posterior-mean estimate $X^*=\mathbb{E}(X_0\mid X_1)$, which is inherently \emph{distortion-oriented}. Finally, as $t \to 0$, the states $X_t$ approach $X_0$ and increasingly concentrate near the clean-speech manifold. In this regime, the stochasticity in $X_t$ is dominated by a small residual Gaussian perturbation around $X_0$, so $X_t$ lies in a narrow neighborhood of the manifold. Hence, the objective mainly enforces local consistency near the data manifold. As a result, the SB model becomes more effective at removing these small perturbations while preserving the distributional statistics of clean speech, which leads to more \emph{perception-oriented} predictions near the end of sampling.

For intermediate time $t \in (0,1)$, predictions become more perceptually realistic as $t$ decreases, yet this gain in perceptual quality is often accompanied by increased point-wise distortion and reduced reconstruction fidelity. Model predictions $X_\theta$ can thus evolve smoothly from \emph{distortion-oriented} to \emph{perception-oriented} estimates, forming a systematic drift trajectory. This smooth evolution introduces an error-induced drift into the sampling trajectory, revealing a practical yet fundamental bias in the sampling direction during SB inference that is not explicitly considered or optimized by the original SB training objective.

To empirically confirm our theoretical analysis, we trained an SB model using the WSJ0+WHAM dataset and examined its 50-step sampling trajectory. Fig.~\ref{fig:pred_error_trajectory}(a) reports the per-step prediction error measured by MSE $\mathbb{E}\!\left[\,\|X_\theta(\cdot)-X_0\|_2^2\,\right]$, which directly reflects the distortion of the model predictions relative to the clean target. The error is low at early steps and gradually accumulates as sampling proceeds, suggesting increasing distortion that peaks near the end of the trajectory. In Fig.~\ref{fig:pred_error_trajectory}~(b), as sampling progresses, the perceptual quality of the model predictions improves as indicated by PESQ, while distortion increases as reflected by a decreasing SI-SDR. This suggests a shift in the model predictions from pursuing point-estimate accuracy to focusing on perceptual quality. These empirical trends validate our theoretical finding that the model prediction $X_\theta$ evolves from a distortion-oriented estimate at $t=1$ toward a perception-oriented sample as $t \to 0$, yielding improved perception at the cost of higher distortion. The detailed experimental setup is shown in Sec.~\ref{sec:experiment_setup}.

\begin{figure}[!t]
\centering
\vspace{-3mm}
\includegraphics[width=0.6 \columnwidth]{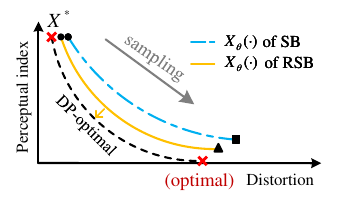}
\vspace{-5mm}
\caption{Illustration of the Distortion-Perception~(DP) tradeoff. A lower perceptual index means better perceptual quality, where lower distortion means higher fidelity. The optimal solutions lie on the DP-optimal frontier. $\blacktriangle$ and $\blacksquare$ respectively represent the model predictions of SB and RSB at the final step, which is also the final predicted clean speech.}
\label{fig:rsb_dp_tradeoff}
\end{figure}

\subsubsection{DP Tradeoff of SB}
Building on the above theoretical and empirical evidence, we observe a systematic drift in the model predictions $X_\theta(\cdot)$ as $t$ decreases. Since $X_\theta$ is algebraically related to the reverse-time score and thus directly determines the reverse drift, this drift propagates into and biases the multi-step inference dynamics. Empirically, the resulting predictions form a curve on the DP plane, which is labeled as ``$X_\theta$ of SB'' in Fig.~\ref{fig:rsb_dp_tradeoff}. Along this curve, improvements in perceptual quality are often accompanied by increased distortion, indicating that the trajectory moves away from the DP-optimal frontier. As a result, the final solution of SB is also not DP-optimal. Accordingly, our goal is to move the final solution closer to the DP-optimal frontier, thereby improving the empirical realism–fidelity tradeoff.

There are two reasons for this sub-optimality. First, the standard SB objective is primarily designed to learn the reverse-time dynamics by matching bridge marginals, rather than to minimize a global discriminative loss that directly penalizes reconstruction error and steers the model toward the minimum-distortion estimate for a given measurement. In other words, such an objective is tradeoff-agnostic and provides no explicit mechanism to reconcile distortion and perception. Second, bridge dynamics implied by the exact potentials are approximated by the SB model, leading to prediction errors. With finite model capacity, the learned dynamics inevitably deviate from the true dynamics, and these errors further compound through iterative sampling, driving the sampling trajectory to drift away and ultimately degrading the overall evaluation of distortion and perception.

\subsubsection{Exposure Bias Problem of SB}
Beyond the DP tradeoff, SB also suffers from exposure bias, i.e., the mismatch between the model inputs seen during training and those encountered during multi-step inference. During training, the SB model is fed the real intermediate states $X_t$, which are sampled from the closed-form marginal $p_t(X_t\mid X_0,X_1)$, so the training inputs follow the ground-truth bridge marginals. During inference, however, each state $\tilde{X}_t$ is sampled using the model prediction $X_\theta(\tilde{X}_\tau,\tau,X_1)$ from the previous step $t=\tau$ via Eq.~(\ref{eq:sb_sde_sampler}). Since the SB model is never exposed to the states corrupted by prediction errors during training, it fails to generalize to states encountered during iterative sampling, which deviate from the training marginals.

Consequently, the inference-time input distribution deviates from the training distribution due to the inevitable prediction errors. This mismatch results in a compounding error effect: Each predicted state deviates from its corresponding ground-truth state, and the resulting error is fed back into subsequent inputs. These errors propagate through the reverse-time dynamics, causing the sampling trajectory to drift away from the ground-truth trajectory. Such error accumulation might ultimately result in a degradation in both perceptual quality and reconstruction fidelity. This is empirically confirmed by results shown in Fig.~\ref{fig:pred_error_trajectory}. However, the standard training objective does not account for this mismatch.

In summary, building on the above analysis, we recognize that the standard SB objective lacks an explicit mechanism to control the DP balance in practice, and it does not account for the train–inference mismatch that leads to error accumulation. On the one hand, the step-wise evolution of model predictions induces a systematic drift, causing the inference trajectory to deviate from the DP-optimal frontier on the DP plane and yielding solutions that are often sub-optimal in terms of the tradeoff. On the other hand, multi-step inference repeatedly conditions on the model’s own imperfect predictions, which progressively enlarges the distribution shift between training and inference and exacerbates exposure bias.

\begin{figure*}[!t]
\centering
\includegraphics[width=1.8 \columnwidth]{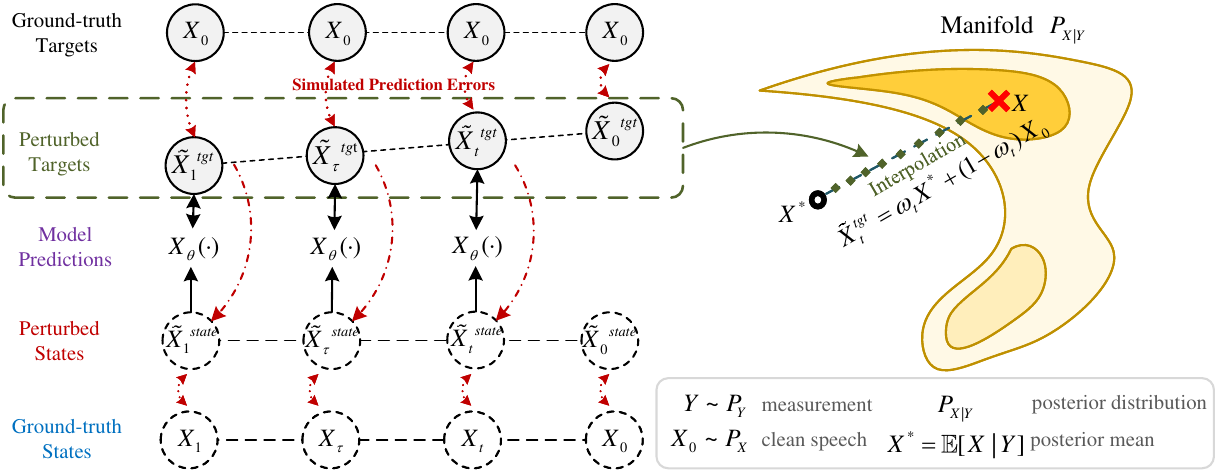}
\vspace{-2mm}
\caption{Illustration of DP Perturbation with an example of a complex posterior $p(X\mid Y)$. DP Perturbation defines time-varying training targets $\tilde{X}^{\mathrm{tgt}}_t$ by interpolating between the clean target $X_0$ and the posterior-mean reference $X^*$, and uses the same interpolation to construct matched perturbed intermediate states $\tilde{X}^{\mathrm{state}}_t$. The posterior-mean reference provides fidelity-preserving guidance by biasing the early-stage transport toward $X^*$, so the trajectory first reaches a proper distortion level before progressively shifting toward the clean endpoint. This joint perturbation reduces the training-inference mismatch, thereby improving multi-step sampling quality.}

\vspace{-5mm}
\label{fig:dp_perturbation_schematic}
\end{figure*}
\subsection{Distortion-Perception Perturbation}
Motivated by our findings, we propose the Distortion-Perception (DP) Perturbation, which is illustrated in Fig.~\ref{fig:dp_perturbation_schematic}, to regularize SB training and address the above two issues.

\subsubsection{Motivation}
Recall that, under multi-step inference, the SB model exhibits a patterned time-dependent drift from distortion-oriented to perception-oriented estimates in its predictions. At early stages (large $t$), the model favors distortion-oriented estimates and often shifts toward the posterior-mean estimate $X^*$. As $t$ decreases along the trajectory, the predictions gradually move toward a clean-speech sample on the data manifold and aim to recover $X_0$.

This behavior is problematic because SB sampling proceeds iteratively. Each estimated clean signal is immediately used to predict the next intermediate state, which feeds the drift into subsequent inputs and compounds step-wise errors over the trajectory. Explicitly simulating such drifts during training exposes the model to its own potential inference-time errors, enabling it to learn robust dynamics that ensure a stable transport toward the clean speech manifold.

\paragraph{Simulating Prediction Drift via Interpolation} To simulate this inference-time drift during training, we introduce a time-varying interpolation that captures how the model prediction shifts from distortion-oriented to perception-oriented behavior along the trajectory. Inspired by~\cite{freirich2021dpt}, under the Wasserstein-2 perception measure, DP-optimal estimators under different perception constraints can be obtained along an interpolation path between a distortion-optimal estimator and a perception-optimal estimator. Each point on the path corresponds to a DP-optimal sample with respect to a specific perception level.

Specifically, we align the interpolation endpoint at $t=1$ with the distortion-optimal reference $X^*$ and the endpoint at $t=0$ with the ground-truth clean signal $X_0$, so that the simulated drift is encouraged to follow the DP-optimal interpolation path. Concretely, we take $X^*=\mathbb{E}[X_0\mid X_1]$ as the distortion-optimal reference, since under the MSE distortion criterion the posterior mean minimizes the expected reconstruction error and is therefore distortion-optimal, and approximate it in practice with a predictive model trained using the MSE loss. This choice is consistent with the empirical behavior shown in Fig.~\ref{fig:pred_error_trajectory}: as sampling proceeds from $t=1$ to $t=0$, the perceptual quality of the model prediction progressively improves, indicating a shift toward higher perceptual realism along the reverse trajectory. Thus, the time-varying interpolation from $X^*$ to $X_0$ provides a dynamic approximation to the DP-optimal interpolation path from the distortion-optimal to the perception-optimal reference.

From the perspective of the DP tradeoff, the perturbed targets provide an explicit fidelity-preserving guide through the distortion-optimal reference $X^*$ at early stages, while gradually shifting back to the clean endpoint as $t \to 0$. This time-dependent supervision encourages the SB model to learn predictions that better balance fidelity and perceptual quality, and drives the learned reverse-time dynamics toward the DP-optimal frontier. From the perspective of exposure bias, training on the perturbed intermediate states exposes the model to error-affected inputs that arise when predictions are reused to advance the sampling trajectory. The model is therefore trained to correct the induced bias, which reduces training–inference mismatch and limits error accumulation.

\paragraph{Relation to Prior Perturbation-Based Methods} Prior perturbation-based training methods~\cite{ning2023ipr, ren2024msds} address the training-inference mismatch of denoising diffusion models by treating prediction errors as Gaussian noise and injecting Gaussian perturbations into intermediate model inputs. These approaches essentially perform robustness training on inputs to increase tolerance to accumulated errors, but they do not alter the underlying diffusion dynamics. In contrast, our DP Perturbation jointly perturbs both the training inputs and the supervision targets. It not only exposes the SB model to inference-like prediction drifts, but also reshapes the transport behavior learned by the SB model. Specifically, DP Perturbation defines a family of time-varying targets that smoothly interpolates between a distortion-optimal endpoint and the perception-optimal endpoint, and pairs each target with matched perturbed intermediate inputs sampled from the induced bridge marginals. As a result, the SB model is trained along a time-varying path motivated by the DP-optimal interpolation principle, while the sampling trajectory is guided to favor the distortion-optimal endpoint at early steps to reduce reconstruction error and then progressively shift toward the perception-optimal endpoint at later steps to recover perceptual realism. By doing so, DP Perturbation actively regularizes the bridge dynamics, enabling a simple yet effective training strategy to mitigate exposure bias and improve the empirical DP tradeoff in a unified manner. This is fundamentally different from prior approaches that passively inject Gaussian noise for robustness.

Furthermore, our perturbation connects to \emph{label smoothing regularization}~\cite{cheng2021srt}. In our work, $X^*$ and $X_0$ can be interpreted as two regression labels that represent distortion-optimal and perception-optimal references. DP Perturbation replaces the original hard label with a time-dependent soft label obtained by smoothly interpolating between these two endpoints, which yields a controlled supervision signal. Hence, DP Perturbation can be viewed as a training regularization.

\subsubsection{Joint Perturbation}
Based on previous analysis, we use the clean sample $X_0$ as the perception-optimal endpoint and the posterior-mean estimate $X^*$ as the distortion-optimal endpoint, defining the interpolation form of our training targets $\tilde{X}_t^{\mathrm{tgt}}$ as,
\begin{equation}
\label{eq:rsb_pert_tgt}
   \tilde{X}_t^{\mathrm{tgt}} = \omega_t X^*+(1-\omega_t) X_0,
\end{equation}
where $\omega_t$ is a predefined interpolation schedule that satisfies $\omega_0 = 0$ and $\omega_1 = 1$. For simplicity, we adopt $\omega_t=t^2$ and keep it fixed across all noise schedules and datasets in our experiments. This quadratic schedule grows slowly for small $t$, so that the targets remain close to $X_0$ as $t\to0$. The resulting deviation of the perturbed target from the clean endpoint is $\| \tilde{X}_t^{\mathrm{tgt}}-X_0 \|_2^2 = \omega_t^2\|X^*-X_0\|_2^2$. This quantity specifies the designed deviation of the supervision target from the clean endpoint, not the empirical prediction-error curve of vanilla SB in Fig.~\ref{fig:pred_error_trajectory}. As shown in Fig.~\ref{fig:dp_perturbation_schematic}, larger target deviations are introduced at larger $t$ to expose the model to early-stage distortion-oriented drift, while smaller $t$ keeps the supervision close to the clean endpoint and encourages progressive correction toward the clean-speech distribution. Additionally, we utilize an auxiliary predictive task to produce $X^*$. A predictive model $D_\phi$ is trained with the MSE training objective. After training, the prediction of $D_\phi$ given $Y$ is considered to approximate $X^*$, i.e., $ X^* \approx D_\phi(Y)$. Notably, $D_\phi$ is a training-only auxiliary network to produce the distortion-optimal reference for DP perturbation.

Following Eq.~(\ref{eq:sb_marginal}), we can obtain the perturbed intermediate states with simulated prediction errors as model inputs,
\begin{equation}
\label{eq:rsb_pert_state}
\tilde{X}_t^{\mathrm{state}} = \mu_x(\tilde{X}_t^{\mathrm{tgt}},X_1,t) + \sigma_x(t) \mathbf{z}, \quad \mathbf{z} \sim \mathcal{N}(0, \mathbf{I}).
\end{equation}

\subsubsection{Regularizing Training with Perturbation}
To regularize model training, we incorporate our perturbations into the training objective. Specifically, we replace the ground-truth training target $X_0$ and the input state $X_t$ in Eq.~(\ref{eq:sb_xpred_loss}) with our perturbed counterparts, $\tilde{X}_t^{\mathrm{tgt}}$(Eq.~(\ref{eq:rsb_pert_tgt})) and $\tilde{X}_t^{\mathrm{state}}$(Eq.~(\ref{eq:rsb_pert_state})), respectively. Hence, our RSB can be formalized with the following regularized training objective,
\begin{equation}
\label{eq:rsb_objective}
\min_\theta \mathbb{E}_{\tilde{X}_t^{\mathrm{state}}, t}[\| X_\theta(\tilde{X}_t^{\mathrm{state}}, t, X_1) - \tilde{X}_t^{\mathrm{tgt}} \|^2_2].
\end{equation}
Algorithm \ref{alg:rsb_training} details the training process, and RSB adopts the same SDE sampler (Eq.~(\ref{eq:sb_sde_sampler})) as the vanilla SB.

When the regularized objective is optimized, RSB is trained on intermediate states with simulated inference-time prediction errors, thereby reducing the training–inference mismatch and mitigating exposure bias. Furthermore, by applying the same time-varying interpolation to both the training targets in Eq.~(\ref{eq:rsb_objective}) and the corresponding input states, RSB provides structured supervision via DP Perturbation that steers the predictions toward distortion-optimal estimates at noisy stages and gradually shifts them back toward the clean endpoint as $t \to 0$. This yields more stable reverse-time dynamics under imperfect intermediate states, limiting error accumulation throughout multi-step sampling. Moreover, the resulting predictions are encouraged to steer toward the DP-optimal frontier, approaching the clean endpoint with high perceptual quality and minimal distortion, so the terminal sample is less likely to drift away from the low-distortion region while still matching the clean-speech manifold. Overall, RSB can produce more faithful and perceptually plausible reconstructions of clean speech.

\begin{algorithm}[t]
\caption{Training procedure of RSB}
\label{alg:rsb_training}
\begin{algorithmic}
\REQUIRE dataset $\mathcal{D}=\{(x,y)\}$, $D_\phi$, $X_\theta$ \\
\STATE \textbf{Stage 1: Predictive model training}
\STATE \quad \textbf{repeat}
\STATE \qquad Draw $(x,y) \sim \mathcal{D}$
\STATE \qquad Update $\phi$ by minimizing $\|D_\phi(y)-x\|_2^2$
\STATE \quad \textbf{until} $D_\phi$ converges
\STATE \textbf{Stage 2: Offline generation of distortion-optimal references}
\STATE \quad Initialize $\mathcal{D}^* \gets \emptyset$
\STATE \quad \textbf{for} each $(x,y)\in\mathcal{D}$ \textbf{do}
\STATE \qquad $x^* \gets D_\phi(y)$ and store $(x,y,x^*)$ in $\mathcal{D}^*$
\STATE \quad \textbf{end for}
\STATE \textbf{Stage 3: RSB training}
\STATE \quad \textbf{repeat}
\STATE \qquad Draw $(x,y,x^*) \sim \mathcal{D}^*$ and $t \sim \mathcal{U}([t_\epsilon, 1])$
\STATE \qquad Obtain perturbed target $\tilde{x}_t^{\mathrm{tgt}}$ via Eq.~\eqref{eq:rsb_pert_tgt}
\STATE \qquad Sample perturbed state $\tilde{x}_t^{\mathrm{state}}$ from Eq.~\eqref{eq:rsb_pert_state}
\STATE \qquad Compute $X_\theta(\tilde{x}_t^{\mathrm{state}}, t, y)$ and update $\theta$ using Eq.~\eqref{eq:rsb_objective}
\STATE \quad \textbf{until} $X_\theta$ converges
\end{algorithmic}
\end{algorithm}

\section{Experimental Setup}\label{sec:experiment_setup}
In this section, we present datasets, evaluation metrics, listening test setup, implementation details, as well as predictive and generative SE methods used for comparison.

\subsection{Datasets}
To compare with prior work on a standard benchmark, we report denoising results on the widely adopted VoiceBank+DEMAND dataset~\cite{valentini2016vbdemand}. We also build large-scale synthetic SE datasets based on the Wall Street Journal (WSJ0) corpus~\cite{garofolo2007wsj0} for evaluating the performance of denoising and dereverberation, which provides clean speech at scale and aligned ground-truth transcripts. All audio recordings are resampled to 16 kHz. Each dataset is described below in detail.

\paragraph{VoiceBank+DEMAND} This is a publicly available benchmark for speech denoising. Each clean utterance from VoiceBank~\cite{veaux2013voicebank} is mixed with one of ten noise recordings, including eight DEMAND recordings~\cite{thiemann2013demand} and two synthetic noises. Mixtures are generated at 15, 10, 5, and 0 dB SNR for training and at 17.5, 12.5, 7.5, and 2.5 dB SNR for testing. The training set contains 10\,802 utterances ($\approx$ 9 hours) from 28 speakers, the validation set 770 utterances, and the test set 824 utterances from 2 unseen speakers.

\paragraph{WSJ0+WHAM} We construct this synthetic speech denoising dataset using clean utterances from the widely used WSJ0 corpus and noise recordings from the WHAM noise corpus~\cite{wichern2019wham}. Each clean utterance is mixed with a randomly selected WHAM noise segment at a signal-to-noise ratio~(SNR) uniformly drawn from -6 dB to 14 dB. The training set uses the \texttt{si\_tr\_s} subset of WSJ0, containing 25\,550 mixtures ($\approx$ 50 h), while the validation and test sets correspond to \texttt{si\_et\_05} and \texttt{si\_dt\_05}, containing 1\,302 and 2\,412 mixtures, respectively. Crucially, the ground-truth transcripts of the test set are preserved to facilitate the performance evaluation of speech recognition.

\paragraph{WSJ0+REVERB} This dataset is created for speech dereverberation with the same data split as WSJ0+WHAM, following the data-generation setting of StoRM~\cite{lemercier2023storm}. The degraded speech is generated by convolving the clean utterances from WSJ0 with synthetic room-impulse responses simulated using the image-source method. Specifically, the target reverberation time $T_{60}$ is uniformly sampled from $[0.4,1.0]$~s, and the room dimensions (length, width, height) are sampled from $[5,15]\times[5,15]\times[2,6]$~m. The source and microphone positions are independently sampled with a minimum distance of $1$~m from the room boundaries. The resulting dataset has an average direct-to-reverberant ratio~(DRR) of around $-9$~dB and an average measured $T_{60}$ of about $0.9$~s. The corresponding anechoic target is generated from a dry room with the same geometry and a fixed absorption coefficient of $0.99$.

\subsection{Evaluation Metrics}
To evaluate the DP tradeoff, we report intrusive metrics and non-intrusive metrics. Intrusive metrics are reference-based, as they require clean references when evaluating enhanced speech. Non-intrusive metrics are reference-free and estimate perceptual quality directly from enhanced speech without access to clean references.

\paragraph{Intrusive Metrics} We first report signal-fidelity metrics, including the Scale-Invariant Signal-to-Distortion Ratio~(SI-SDR)~\cite{leroux2019sdr}, Scale-Invariant Signal-to-Interference Ratio~(SI-SIR), and Scale-Invariant Signal-to-Artifact Ratio~(SI-SAR), which measure overall distortion, interference, and algorithmic artifacts, respectively. These metrics directly reflect signal fidelity to the clean reference and correspond to the distortion dimension of the DP tradeoff. We also report intrusive metrics for perceptual quality and intelligibility, including the Perceptual Evaluation of Speech Quality~(PESQ)~\cite{rix2001pesq}, the composite objective quality metrics~\cite{hu2008eoqm} (CSIG for signal distortion, CBAK for background intrusiveness, and COVL for overall quality), and Extended Short-Time Objective Intelligibility~(ESTOI)~\cite{jensen2016estoi}. Although these metrics are designed to approximate human perception or intelligibility, their reference-based formulations evaluate perceptual or intelligibility-related similarity relative to a specific clean reference and may therefore not fully capture reference-free perceptual naturalness.

\paragraph{Non-Intrusive Metrics} We report DNSMOS P.808~(referred to as DNSMOS)~\cite{reddy2021dnsmos}, UTMOS~\cite{saeki2022utmos}, and the overall score of SIGMOS~(referred to as SIGMOS)~\cite{ristea2025ssi}. These metrics estimate perceptual quality directly from enhanced speech, providing complementary evidence of speech naturalness. DNSMOS was trained on ratings collected in the DNS Challenge and is closely tied to SE conditions involving residual noise and noise-suppression artifacts~\cite{reddy2021dnsmos}. In contrast, UTMOS and SIGMOS were trained for MOS prediction from broader listener rating corpora, and therefore better reflect human preference~\cite{saeki2022utmos,ristea2025ssi}.

Meanwhile, we validate the linguistic preservation of RSB via a downstream automatic speech recognition (ASR) evaluation. For WSJ0+WHAM and WSJ0+REVERB, the enhanced utterances produced by each method are transcribed using \emph{Whisper base.en}~\cite{radford2022whisper} (an English-only speech recognition model with approximately 74M parameters). We compute utterance-level word error rates (WERs) with the Python library \textit{jiwer} and report the mean WER over all test utterances.

\subsection{Listening Test Setup}
To evaluate the subjective performance of RSB, we conducted a medium-scale subjective listening test using webMUSHRA~\cite{schoeffler2018webmushra}, with 12 participants. Ten samples were selected from the WSJ0+WHAM and WSJ0+REVERB datasets for perceptual audio quality evaluation. In each trial of the test, participants were presented with a reference audio and several audio samples in random order, and asked to assign ratings between 1 and 100 to measure the perceptual audio quality of each sample.

\begin{table}[!t]
    \centering
    \caption{Noise schedules for both SB and RSB.}
    \label{tab:schedules_vp_ve}
    \setlength{\tabcolsep}{6pt}
    \renewcommand{\arraystretch}{1.15}
    \begin{tabular}{l|c|c}
        \toprule
        Schedule& VE & VP \\
        \midrule
        $f(t)$
        & $0$
        & $-\dfrac{1}{2}\big[\beta_0+(\beta_1-\beta_0)t\big]$ \\
        \midrule
        $g^2(t)$
        & $c k^{2t}$
        & $c\big[\beta_0+(\beta_1-\beta_0)t\big]$ \\
        \midrule
        $\alpha_t$
        & $1$
        & $e^{-\frac{1}{2}(\beta_0 t+\frac{\beta_1-\beta_0}{2}t^2)}$ \\
        \midrule
        $\sigma_t^2$
        & $\dfrac{c\,(k^{2t}-1)}{2\log k}$
        & $c(e^{\beta_0 t+\frac{\beta_1-\beta_0}{2}t^2}-1)$ \\
        \bottomrule
    \end{tabular}
\end{table}

\subsection{Implementation Details}
\subsubsection{Data Representation} The complex spectrogram features are extracted using the short-time Fourier transform (STFT)~\cite{lemercier2023storm}. We use a square-root Hann window with a window length of 510 and a hop length of 128, and apply a square-root magnitude warping operation as in~\cite{richter2023sgmseplus}. During training, the original utterances are randomly truncated into sequences of 256 STFT frames ($\approx$ $2~s$) and then normalized by the maximum absolute value of the corresponding measurement signal.

\subsubsection{Noise Schedule} Detailed schedules are shown in Table~\ref{tab:schedules_vp_ve}. Table~\ref{tab:schedules_vp_ve} reports the concrete VP and VE schedule instantiations used in our experiments under the tractable SB parameterization introduced in Eq.~(\ref{eq:sb_solution}).
Unless stated otherwise, we use the VE schedule as the default for both SB and RSB, with $c=0.40$ and $k=2.6$ following~\cite{jukic2024sbse}. The VP schedule is also considered in this paper, using $\beta_0=0.01$, $\beta_1=20$, and $c=0.3$, following~\cite{jukic2024sbse}.

\subsubsection{Training Details}
The SB baseline and RSB are implemented by us following the tractable SB formulation~\cite{jukic2024sbse}. We adopt NCSN++M as the default architecture for $D_\phi$ because using the same NCSN++M architecture for $D_\phi$ and the SB model $X_\theta$ matches their model capacity and aligns the prediction behavior of $D_\phi$ with the SB dynamics. The resulting distortion-optimal references are therefore better suited to simulating inference-time prediction drift. This architectural choice also follows the practice adopted in StoRM~\cite{lemercier2023storm}. NCSN++M has approximately 25.2 M parameters~\cite{richter2023sgmseplus, lemercier2023storm}.

The loss in Eq.~(\ref{eq:rsb_objective}) is computed on the complex spectrogram, and we additionally include a time-domain reconstruction loss following~\cite{jukic2024sbse}. All models are trained on a single NVIDIA A100 GPU. The minimum time $t_\epsilon$ during training is set to $10^{-4}$.

During training, the batch size is set to 16, and we use the Adam optimizer~\cite{kingma2015adam} with a learning rate of $10^{-4}$.
In addition, we train our models for up to 1000 epochs with early stopping based on the average SI-SDR computed on 50 validation utterances, with a patience of 50 epochs. To stabilize training and improve generalization, we maintain an exponential moving average (EMA) of the model parameters $\theta$ with a decay factor of 0.999, and select the EMA checkpoint that achieves the best PESQ on the validation set as the final checkpoint~\cite{jukic2024sbse}.

The overall training procedure is shown in Algorithm~\ref{alg:rsb_training}. First, $D_\phi$ is trained with the MSE objective. Second, after $D_\phi$ has been fully trained, we run it once offline on the training set to compute the distortion-optimal references $X^*$ required by DP Perturbation, and save them to disk. Third, the SB model $X_\theta$ is trained by loading the cached references directly from disk. On WSJ0+WHAM and WSJ0+REVERB, training each of $D_\phi$ and $X_\theta$ takes approximately $40$~hours. On VoiceBank+DEMAND, each takes approximately $10$~hours due to its smaller training set. Because $D_\phi$ is not used at inference and the distortion-optimal references are cached before RSB training, RSB incurs no additional online evaluations of $D_\phi$ during its training iterations. The source code and audio samples are available online\footnote{\url{https://yorch233.github.io/RSB/}}.

\subsection{Compared Methods}
We compare RSB with both advanced predictive and diffusion-based generative methods. On VoiceBank+DEMAND, we expand the evaluation to include purely predictive NCSN++M and several advanced predictive SE methods, including TF-GridNet~\cite{wang2022tfgridnet}, MetricGAN+~\cite{fu2021metricganplus}, SEMamba~\cite{chao2024semamba}, and MP-SENet~\cite{lu2023mpsenet}. We select this public and widely adopted benchmark because most of these methods have reported results and released pretrained checkpoints on it, enabling a fair and reproducible comparison on the same test set. NCSN++M is the default architecture of $D_\phi$ used to construct $X^*$ in our framework, and its reported results represent the performance of this distortion-optimal reference. We also evaluate diffusion-based generative baselines, including CDiffuSE~\cite{lu2022cdiffuse}, SGMSE+~\cite{richter2023sgmseplus}, StoRM~\cite{lemercier2023storm}, and SB~\cite{jukic2024sbse}. On the larger and more challenging WSJ0+WHAM and WSJ0+REVERB datasets, we retain the original comparison set for an in-depth evaluation of diffusion-based generative methods across denoising and dereverberation.

The SB baseline and RSB are based on our own implementation, and use the same SDE sampler described in Eq.~(\ref{eq:sb_sde_sampler}). For the other methods, on VoiceBank+DEMAND, we use officially released checkpoints whenever available. Since TF-GridNet does not provide a directly applicable checkpoint for this benchmark, we train an efficient publicly available third-party implementation\footnote{\url{https://github.com/philgzl/brever}} on VoiceBank+DEMAND. On WSJ0+WHAM and WSJ0+REVERB, we retrain the evaluated baselines using the official code released with the corresponding papers. For inference, all diffusion-based models use 50 sampling steps. Since SGMSE+ and StoRM employ a predictor-corrector sampler~\cite{richter2023sgmseplus,lemercier2023storm}, this setting corresponds to 100 network calls for these two methods. It is also worth noting that StoRM, SB, and RSB all adopt NCSN++M as the default model architecture.

To determine whether the performance differences between SB and RSB are statistically significant, we conduct paired $t$-tests on utterance-level metric scores by pairing the outputs of the two methods for each test utterance. We adopt a significance threshold of $p<0.005$ for all comparisons.

\section{Experimental Results and Discussion}\label{sec:experiment_results}
\subsection{Quantitative Comparison}
\begin{table*}[!t]
    \centering
    \caption{Denoising performance on the VoiceBank+DEMAND dataset. Results are reported as mean~$\pm$~std. For each metric, the best result within each method group is highlighted in bold, and the second-best is underlined.}
    \label{tab:vb_main}
    \setlength{\tabcolsep}{4pt}
    \resizebox{\textwidth}{!}{%
    \begin{tabular}{@{}lc|ccccc|ccc@{}}
        \toprule
        Method & Type & PESQ$\uparrow$ & ESTOI$\uparrow$ & CSIG$\uparrow$ & CBAK$\uparrow$ & COVL$\uparrow$ & SI-SDR$\uparrow$ & SI-SIR$\uparrow$ & SI-SAR$\uparrow$ \\
        \midrule
        unprocessed & -- & 1.97~$\pm$~0.75 & 0.79~$\pm$~0.15 & 3.34~$\pm$~0.87 & 2.44~$\pm$~0.66 & 2.63~$\pm$~0.83 & 8.5~$\pm$~5.6 & 8.5~$\pm$~5.6 & -  \\
        \midrule
        NCSN++M     & P  & 2.82~$\pm$~0.72 & \underline{0.87}~$\pm$~0.10 & 3.69~$\pm$~0.89 & 3.43~$\pm$~0.58 & 3.24~$\pm$~0.83 & \underline{19.6}~$\pm$~3.6 & \textbf{32.6}~$\pm$~7.5 & 20.1~$\pm$~3.6 \\
        TF-GridNet~\cite{wang2022tfgridnet} & P & 2.86~$\pm$~0.58 & \underline{0.87}~$\pm$~0.09 & 4.27~$\pm$~0.54 & 3.49~$\pm$~0.49 & 3.58~$\pm$~0.58 & 19.2~$\pm$~3.6 & 27.8~$\pm$~5.1 & \underline{20.4}~$\pm$~3.7 \\
        MetricGAN+~\cite{fu2021metricganplus} & P & 3.13~$\pm$~0.55 & 0.83~$\pm$~0.11 & 4.12~$\pm$~0.68 & 3.16~$\pm$~0.45 & 3.62~$\pm$~0.65 & 8.5~$\pm$~3.8 & 26.6~$\pm$~4.2 & 8.6~$\pm$~3.8 \\
        SEMamba~\cite{chao2024semamba} & P & \underline{3.54}~$\pm$~0.61 & \textbf{0.89}~$\pm$~0.08 & \underline{4.70}~$\pm$~0.45 & \underline{3.87}~$\pm$~0.48 & \underline{4.20}~$\pm$~0.59 & \textbf{19.7}~$\pm$~3.2 & \underline{30.4}~$\pm$~5.7 & \textbf{20.5}~$\pm$~3.3 \\
        MP-SENet~\cite{lu2023mpsenet} & P & \textbf{3.61}~$\pm$~0.54 & \textbf{0.89}~$\pm$~0.08 & \textbf{4.73}~$\pm$~0.39 & \textbf{3.89}~$\pm$~0.46 & \textbf{4.24}~$\pm$~0.52 & 19.4~$\pm$~3.5 & 29.4~$\pm$~4.5 & 20.2~$\pm$~3.8 \\
        \midrule
        CDiffuSE~\cite{lu2022cdiffuse}       & G & 2.52~$\pm$~0.61 & 0.80~$\pm$~0.11 & 3.73~$\pm$~0.64 & 2.93~$\pm$~0.48 & 3.10~$\pm$~0.64 & 13.5~$\pm$~2.4 & 19.2~$\pm$~4.9 & 15.5~$\pm$~1.5 \\
        SGMSE+~\cite{richter2023sgmseplus}      & G & 2.92~$\pm$~0.64 & \underline{0.86}~$\pm$~0.10 & \underline{4.13}~$\pm$~0.67 & 3.38~$\pm$~0.50 & 3.53~$\pm$~0.67 & 17.5~$\pm$~3.2 & 29.3~$\pm$~5.7 & 18.1~$\pm$~3.2 \\
        StoRM~\cite{lemercier2023storm}      & G & \underline{2.93}~$\pm$~0.66 & \textbf{0.87}~$\pm$~0.10 & \textbf{4.18}~$\pm$~0.66 & \underline{3.47}~$\pm$~0.52 & \underline{3.56}~$\pm$~0.68 & \underline{18.6}~$\pm$~3.4 & \underline{30.9}~$\pm$~6.8 & \underline{19.1}~$\pm$~3.4 \\
        SB~\cite{jukic2024sbse}              & G & \textbf{3.04}~$\pm$~0.69 & \textbf{0.87}~$\pm$~0.09 & 4.06~$\pm$~0.71 & 3.45~$\pm$~0.53 & \underline{3.56}~$\pm$~0.72 & 17.8~$\pm$~3.4 & 30.3~$\pm$~7.4 & 18.5~$\pm$~3.3 \\
        RSB (ours)                           & G & \textbf{3.04}~$\pm$~0.67 & \textbf{0.87}~$\pm$~0.09 & 4.09~$\pm$~0.70\textsuperscript{*} & \textbf{3.50}~$\pm$~0.53\textsuperscript{*} & \textbf{3.57}~$\pm$~0.70 & \textbf{18.8}~$\pm$~3.5\textsuperscript{*} & \textbf{31.2}~$\pm$~6.6\textsuperscript{*} & \textbf{19.4}~$\pm$~3.5\textsuperscript{*} \\
        \bottomrule
    \end{tabular}
    }
    \vspace{0.5ex}
    \parbox{\textwidth}{\footnotesize\raggedright\textit{Note:} \textsuperscript{*} marks the metric scores of RSB, which are significantly better than those of SB under a paired $t$-test ($p<0.005$).}
    
\end{table*}

\begin{table}[!t]
    \centering
    \caption{Non-intrusive perceptual quality metric scores on the VoiceBank+DEMAND dataset. Results are reported as mean~$\pm$~std. For each metric, the best result within each method group is highlighted in bold, and the second-best is underlined.}
    \label{tab:vb_mos_scores}
    \footnotesize
    \begin{tabular}{@{}lc|ccc@{}}
        \toprule
        Method & Type & DNSMOS$\uparrow$ & UTMOS$\uparrow$ & SIGMOS$\uparrow$ \\
        \midrule
        unprocessed & -- & 3.05~$\pm$~0.39 & 2.84~$\pm$~0.61 & 2.56~$\pm$~0.50 \\
        \midrule
        NCSN++M     & P  & \underline{3.56}~$\pm$~0.30 & 3.42~$\pm$~0.45 & \textbf{3.44}~$\pm$~0.33 \\
        TF-GridNet~\cite{wang2022tfgridnet} & P & 3.50~$\pm$~0.29 & \underline{3.48}~$\pm$~0.42 & 3.29~$\pm$~0.31 \\
        MetricGAN+~\cite{fu2021metricganplus} & P & 3.37~$\pm$~0.30 & 3.05~$\pm$~0.51 & 3.22~$\pm$~0.37 \\
        SEMamba~\cite{chao2024semamba} & P & \textbf{3.57}~$\pm$~0.29 & \textbf{3.52}~$\pm$~0.43 & 3.38~$\pm$~0.29 \\
        MP-SENet~\cite{lu2023mpsenet} & P & \textbf{3.57}~$\pm$~0.28 & \textbf{3.52}~$\pm$~0.42 & \underline{3.40}~$\pm$~0.29 \\
        \midrule
        CDiffuSE~\cite{lu2022cdiffuse}       & G & 3.13~$\pm$~0.34 & 3.08~$\pm$~0.50 & 2.87~$\pm$~0.40 \\
        SGMSE+~\cite{richter2023sgmseplus}      & G & 3.53~$\pm$~0.29 & \textbf{3.68}~$\pm$~0.41 & \underline{3.39}~$\pm$~0.30 \\
        StoRM~\cite{lemercier2023storm}      & G & \textbf{3.57}~$\pm$~0.32 & 3.54~$\pm$~0.41 & 3.37~$\pm$~0.31 \\
        SB~\cite{jukic2024sbse}              & G & \textbf{3.57}~$\pm$~0.28 & 3.61~$\pm$~0.42 & \textbf{3.51}~$\pm$~0.31 \\
        RSB (ours)                           & G & \underline{3.56}~$\pm$~0.29 & \underline{3.64}~$\pm$~0.41 & \textbf{3.51}~$\pm$~0.32 \\
        \bottomrule
    \end{tabular}
    
\end{table}

\paragraph{Denoising Performance on VoiceBank+DEMAND} We first evaluate RSB on the widely used VoiceBank+DEMAND benchmark, comparing it against advanced predictive and diffusion-based SE methods. Table~\ref{tab:vb_main} reports intrusive metrics for quality, intelligibility, and signal fidelity, while Table~\ref{tab:vb_mos_scores} provides complementary non-intrusive perceptual quality metrics. In these tables, P and G denote predictive and generative methods, respectively.

Among the predictive SE methods, MP-SENet achieves the highest PESQ and composite quality scores and ties SEMamba for the highest ESTOI, while SEMamba attains the highest SI-SDR and SI-SAR and NCSN++M attains the highest SI-SIR. These results reflect the strong reference-aligned reconstruction and high signal fidelity achieved by supervised discriminative training. MetricGAN+, SEMamba, and MP-SENet further incorporate metric-oriented optimization guided by PESQ, which can directly favor performance on this metric~\cite{fu2021metricganplus,chao2024semamba,lu2023mpsenet}. However, this advantage does not consistently extend to the non-intrusive metrics~\cite{richter2025ito}. DNSMOS assigns comparable scores to the strongest predictive and generative methods, whereas SGMSE+, RSB, and SB achieve higher UTMOS scores than all predictive baselines, and SB and RSB achieve the highest SIGMOS scores. This contrast indicates that the predictive methods provide stronger reference-aligned fidelity but slightly lower estimated speech naturalness than the strongest generative methods.

Among the generative approaches, RSB exhibits a more favorable fidelity-realism balance than both the score-based diffusion SE method SGMSE+~\cite{richter2023sgmseplus} and the hybrid SE method StoRM~\cite{lemercier2023storm}. Its higher SI-SDR, together with improvements in PESQ and SIGMOS over SGMSE+, indicates that the gain in reconstruction fidelity is not obtained at the expense of perceptual speech quality. Similarly, its advantages over StoRM in SI-SDR, PESQ, UTMOS, and SIGMOS suggest that a strong fidelity-realism balance can be achieved without relying on an auxiliary predictive model in a cascaded framework.

Compared with SB, RSB improves SI-SDR by 1.0~dB, SI-SIR by 0.9~dB, and SI-SAR by 0.9~dB, indicating enhanced speech reconstruction, interference suppression, and artifact control, respectively. Together with the gains in CSIG and CBAK, all five improvements are statistically significant. The CSIG and CBAK results further support improved estimated speech quality and reduced background-noise intrusiveness. By contrast, PESQ and ESTOI remain unchanged at the reported precision, while the non-intrusive metrics remain comparable, showing that the fidelity improvements are achieved without an evident degradation in perceptual quality. The SI-SDR gap to SEMamba is consequently reduced from 1.9~dB for SB to 0.9~dB for RSB. These results provide statistically supported evidence that the proposed RSB improves the speech fidelity of the standard SB formulation while preserving its perceptual quality.

\begin{table*}[!t]
\centering
\caption{Performance Comparison on the WSJ0+WHAM and WSJ0+REVERB datasets. Results are reported as mean~$\pm$~std. For each metric, the best result is highlighted in bold, and the second-best is underlined.}
\label{tab:wsj_main_results}
\setlength{\tabcolsep}{4pt}
\resizebox{\textwidth}{!}{%
\begin{tabular}{@{}l|cccccc|ccc@{}}
    \toprule
    Method & PESQ$\uparrow$ & ESTOI$\uparrow$ & CSIG$\uparrow$ & CBAK$\uparrow$ & COVL$\uparrow$ & DNSMOS$\uparrow$ & SI-SDR$\uparrow$ & SI-SIR$\uparrow$ & SI-SAR$\uparrow$\\
    \midrule
    \multicolumn{10}{c}{\textbf{WSJ0+WHAM (Denoising)}}\\
    \midrule
            unprocessed                           & 1.32~$\pm$~0.30              &0.55~$\pm$~0.17& 2.78~$\pm$~0.67              & 1.91~$\pm$~0.50               & 1.99~$\pm$~0.49                      & 2.71~$\pm$~0.30          &  3.9~$\pm$~5.8           &  3.9~$\pm$~5.8           &  –                \\
            \midrule
            NCSN++M                               & 2.18~$\pm$~0.62              &\textbf{0.78}~$\pm$~0.14& 3.05~$\pm$~0.80              & 2.97~$\pm$~0.56               & 2.58~$\pm$~0.74                      & \textbf{3.77}~$\pm$~0.35 & \textbf{14.7}~$\pm$~4.2  & \textbf{29.5}~$\pm$~4.4 & \textbf{14.9}~$\pm$~4.3 \\
            SGMSE+~\cite{richter2023sgmseplus}       & 2.14~$\pm$~0.72              &0.71~$\pm$~0.17& 3.71~$\pm$~0.74              & 2.73~$\pm$~0.67               & 2.90~$\pm$~0.75                     & 3.44~$\pm$~0.39          & 10.3~$\pm$~5.9           & 21.7~$\pm$~9.1          & 11.1~$\pm$~5.3         \\
            StoRM~\cite{lemercier2023storm}       & 2.53~$\pm$~0.60              &0.76~$\pm$~0.14& \textbf{4.06}~$\pm$~0.60     & 3.11~$\pm$~0.51               & \underline{3.29}~$\pm$~0.61         & 3.65~$\pm$~0.38          & 13.4~$\pm$~4.1           & 28.2~$\pm$~4.5          & 13.6~$\pm$~4.1                   \\
            SB~\cite{jukic2024sbse}               & \underline{2.54}~$\pm$~0.62  &\underline{0.77}~$\pm$~0.14& 3.93~$\pm$~0.62              & \underline{3.12}~$\pm$~0.53   & 3.22~$\pm$~0.62                     & \underline{3.74}~$\pm$~0.39          & 13.2~$\pm$~4.3           & \underline{29.2}~$\pm$~4.6          & 13.4~$\pm$~4.4         \\
            RSB (ours)                            & \textbf{2.61}~$\pm$~0.60\textsuperscript{*}     &\underline{0.77}~$\pm$~0.14\textsuperscript{*}& \underline{4.00}~$\pm$~0.59\textsuperscript{*}  & \textbf{3.18}~$\pm$~0.51\textsuperscript{*} & \textbf{3.30}~$\pm$~0.60\textsuperscript{*}  & 3.73~$\pm$~0.39  & \underline{13.7}~$\pm$~4.1\textsuperscript{*}           & \textbf{29.5}~$\pm$~4.6\textsuperscript{*} & \underline{13.8}~$\pm$~4.2\textsuperscript{*}         \\
    \midrule
    \multicolumn{10}{c}{\textbf{WSJ0+REVERB (Dereverberation)}}\\
    \midrule
            unprocessed     & 1.38~$\pm$~0.26         &0.42~$\pm$~0.13& 2.93~$\pm$~0.41         & 1.72~$\pm$~0.28         & 2.09~$\pm$~0.35         & 2.88~$\pm$~0.27          &  -9.1~$\pm$~6.6          & -9.1~$\pm$~6.6           & –                 \\
            \midrule
            NCSN++M         & 1.96~$\pm$~0.47         &0.76~$\pm$~0.11& 3.29~$\pm$~0.53         & 2.55~$\pm$~0.39         & 2.59~$\pm$~0.51         & 3.52~$\pm$~0.37          &   \underline{4.8}~$\pm$~4.4          & 23.6~$\pm$~9.5           &  \underline{5.0}~$\pm$~4.4         \\
            SGMSE+~\cite{richter2023sgmseplus}       & 2.22~$\pm$~0.52             &0.71~$\pm$~0.13     & 3.93~$\pm$~0.43              & 2.52~$\pm$~0.44              & 3.06~$\pm$~0.49                     & 3.56~$\pm$~0.44          & -3.2~$\pm$~8.7           & 5.7~$\pm$~14.1          & 1.4~$\pm$~4.3         \\
            StoRM~\cite{lemercier2023storm}        & 2.52~$\pm$~0.49         &0.75~$\pm$~0.12& \textbf{4.19}~$\pm$~0.37         & 2.86~$\pm$~0.36         & \underline{3.36}~$\pm$~0.44         & \underline{3.63}~$\pm$~0.39          & 4.4~$\pm$~4.5           & \underline{25.8}~$\pm$~10.1          & 4.6~$\pm$~4.1   \\
            SB~\cite{jukic2024sbse}           & \underline{2.58}~$\pm$~0.49         &\underline{0.78}~$\pm$~0.11& \underline{4.18}~$\pm$~0.39         & \underline{2.92}~$\pm$~0.37         & \textbf{3.38}~$\pm$~0.45         & \textbf{3.71}~$\pm$~0.39          &   4.6~$\pm$~4.7          & 25.1~$\pm$~10.4          &  4.7~$\pm$~4.4         \\
            RSB (ours)      & \textbf{2.59}~$\pm$~0.49 &\textbf{0.79}~$\pm$~0.11\textsuperscript{*}& \underline{4.18}~$\pm$~0.39 & \textbf{2.94}~$\pm$~0.38\textsuperscript{*} & \textbf{3.38}~$\pm$~0.45 & \textbf{3.71}~$\pm$~0.39 & \textbf{5.2}~$\pm$~4.5\textsuperscript{*}  & \textbf{26.0}~$\pm$~10.1\textsuperscript{*}          &  \textbf{5.3}~$\pm$~4.3\textsuperscript{*}         \\
    \bottomrule
\end{tabular}
}
\vspace{0.5ex}

\parbox{\textwidth}{\footnotesize\raggedright\textit{Note:} \textsuperscript{*} marks the metric scores of RSB, which are significantly better than those of SB under a paired $t$-test ($p<0.005$).}
\end{table*}

\paragraph{Denoising Performance on WSJ0+WHAM} To further evaluate the effectiveness of our proposed method in speech denoising, we conduct experiments on the WSJ0+WHAM dataset to compare RSB with predictive NCSN++M and advanced generative methods. As shown in Table~\ref{tab:wsj_main_results}, RSB obtains the best or second-best scores across most perception-oriented metrics and achieves the best distortion performance among all diffusion-based methods. Although NCSN++M maintains a slight advantage in DNSMOS and achieves the highest SI-SDR, this comes at the cost of lower perceptual quality (e.g., a PESQ score 0.43 lower than that of RSB). StoRM, with its hybrid framework, outperforms SGMSE+, but performs comparably to SB on WSJ0+WHAM. Notably, compared with SB, RSB achieves significant performance gains in PESQ, ESTOI, CSIG, CBAK, COVL, SI-SDR, SI-SIR, and SI-SAR, while DNSMOS remains comparable. The simultaneous SI-SIR and SI-SAR gains over SB further indicate stronger suppression of residual interference without increasing enhancement artifacts.

\paragraph{Dereverberation Performance on WSJ0+REVERB} We assess the robustness of RSB in reverberant environments and compare its dereverberation performance with baselines on the WSJ0+REVERB dataset in Table~\ref{tab:wsj_main_results}. RSB achieves the best or second-best scores across all perceptual quality metrics and the best performance on all signal-fidelity metrics. Compared with SB, RSB significantly improves ESTOI, CBAK, SI-SDR, SI-SIR, and SI-SAR, while maintaining comparable PESQ, CSIG, COVL, and DNSMOS. RSB also numerically surpasses the predictive NCSN++M baseline on all three signal-fidelity metrics. These results extend the favorable fidelity-realism balance observed for additive denoising to a distinct convolutional degradation characterized by temporally extended room reflections.

\begin{figure*}[!t]
\centering
\vspace{-3mm}
\includegraphics[width=2 \columnwidth]{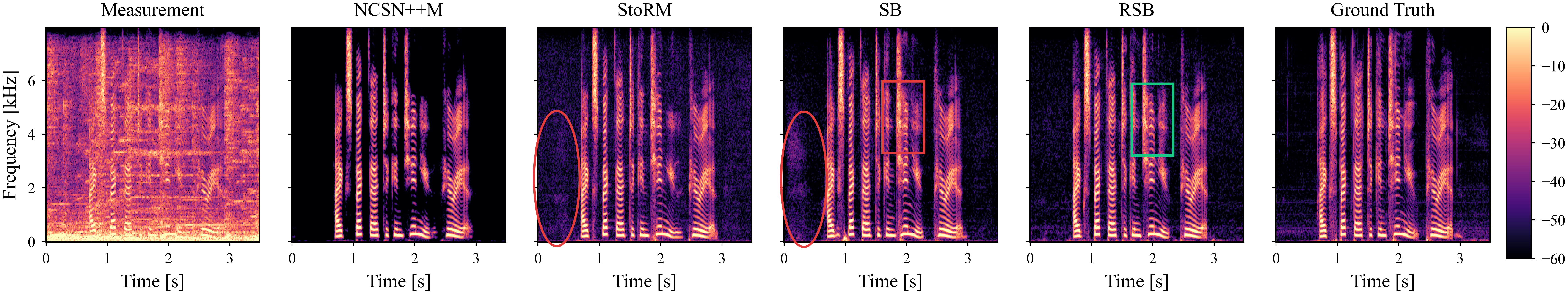}\\[-1mm]
{\small (a) WSJ0+WHAM}
\begin{tikzpicture}[baseline=(reverb.south)]
\node[inner sep=0] (reverb) {\includegraphics[width=2 \columnwidth]{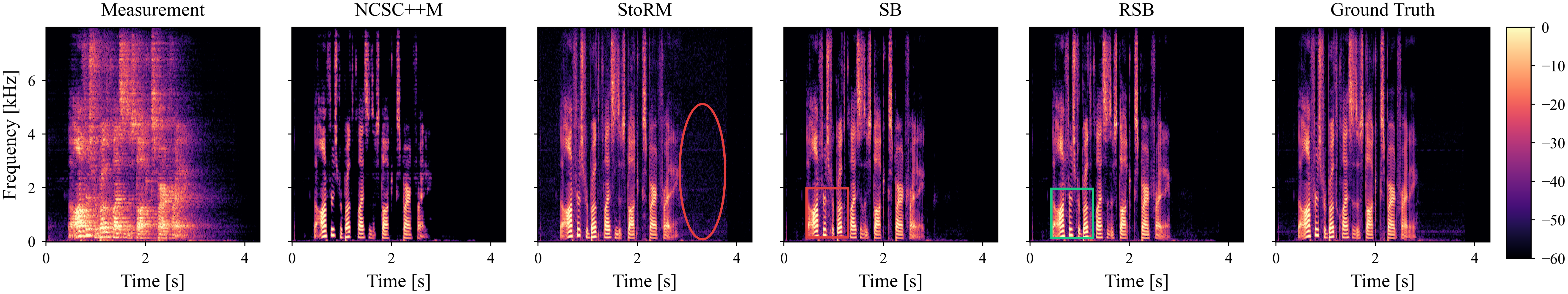}};
\node[anchor=north, fill=white, inner sep=1pt, font=\scriptsize] at ([xshift=-0.49\columnwidth]reverb.north) {NCSN++M};
\end{tikzpicture}\\[-1mm]
{\small (b) WSJ0+REVERB}
\vspace{-2mm}
\caption{Visualization of unprocessed and enhanced utterances. (a): The measurement speech from the WSJ0+WHAM dataset is generated with additive noise at $\mathrm{SNR}=-5.9$ dB. (b): The measurement speech from the WSJ0+REVERB dataset has a measured reverberation time of $T_{60}=1.85$~s.}
\vspace{-3mm}
\label{fig:spectrogram_visualization}
\end{figure*}

\subsection{Qualitative Comparison}
To complement the quantitative results, we present spectrogram visualizations of two test utterances from WSJ0+WHAM and WSJ0+REVERB in Fig.~\ref{fig:spectrogram_visualization}. This qualitative analysis helps interpret how different methods shape harmonic structures and residual artifacts. In summary, NCSN++M produces an overly-suppressed spectrogram with reduced high-frequency content and less distinct harmonic structures, which may affect speech clarity and perceived quality. StoRM and SB show noticeable residual background components, whereas RSB yields a more faithful reconstruction of harmonic patterns and improves spectral fidelity over SB.

\subsection{Ablation Study}
\begin{table}[t]
    \centering
    \caption{Denoising Performance on the VoiceBank+DEMAND dataset using different perturbation strategies. The three strategies are reported under both SDE and ODE samplers.}
    \label{tab:ablation_perturbation}
    \footnotesize
    \setlength{\tabcolsep}{2.8pt}
    \renewcommand{\arraystretch}{1.08}
    \begin{tabular}{@{}llccccc@{}}
        \toprule
        Strategy & Sampler & PESQ$\uparrow$ & CSIG$\uparrow$ & CBAK$\uparrow$ & COVL$\uparrow$ & SI-SDR$\uparrow$ \\
        \midrule
        \multirow{2}{*}{No} & SDE & \textbf{3.04} & 4.06 & 3.45 & 3.56 & 17.8 \\
                            & ODE & 2.68 & 3.39 & 3.12 & 3.04 & 14.7 \\
        \midrule
        \multirow{2}{*}{Input-only} & SDE & 2.98 & 4.02 & 3.44 & 3.50 & 18.1 \\
                                      & ODE & 2.93 & 3.68 & 3.34 & 3.31 & 16.8 \\
        \midrule
        \multirow{2}{*}{Joint} & SDE & \textbf{3.04} & \textbf{4.09} & \textbf{3.50} & \textbf{3.57} & \textbf{18.8} \\
                            & ODE & \textbf{3.01} & \textbf{3.82} & \textbf{3.41} & \textbf{3.43} & \textbf{17.1} \\
        \bottomrule
    \end{tabular}
\end{table}
We conduct ablation studies on the VoiceBank+DEMAND dataset to examine the role of joint perturbation in DP Perturbation and validate the consistent effectiveness of DP Perturbation across various noise schedules.

\paragraph{Perturbation Strategy} We consider three training configurations for comparison. ``No'' denotes vanilla SB without perturbation, ``Input-only'' perturbs only the intermediate states fed to the SB model, and ``Joint'' applies the perturbation consistently to both inputs and targets.

As shown in Table~\ref{tab:ablation_perturbation}, considering both samplers, ``Joint'' achieves the best overall performance, confirming the advantage of aligning perturbed inputs with matched perturbed targets. ODE sampling generally leads to lower scores than SDE sampling, since the deterministic trajectory removes the stochastic correction in the SDE sampler and becomes more sensitive to accumulated prediction errors. For ``No'', this drop is more pronounced because the vanilla SB model is not exposed to error-affected intermediate states during training. ``Input-only'' alleviates this degradation by perturbing inputs, but the unchanged clean targets still introduce a supervision mismatch and slightly degrade perception-oriented metrics. In contrast, ``Joint'' consistently improves almost all metrics, and maintains the best ODE performance among the three strategies. Overall, this ablation validates the design choice of joint perturbation as a fundamental component of DP Perturbation, rather than a standalone robustness technique.

\begin{table}[t!]
    \centering
    \caption{Denoising performance of RSB on the VoiceBank+DEMAND dataset under VE and VP noise schedules.}
    \label{tab:ablation_noise_schedule}
    \footnotesize
    \begin{tabular}{@{}c|c|cccc|c@{}}
        \toprule
        Schedule   &Method    &PESQ$\uparrow$  &CSIG$\uparrow$  &CBAK$\uparrow$  &COVL$\uparrow$  & SI-SDR$\uparrow$  \\
        \midrule
        \multirow{2}{*}{VE} & SB& \textbf{3.04} & 4.06 & 3.45 & 3.56 & 17.8 \\
        &RSB& \textbf{3.04} & \textbf{4.09} & \textbf{3.50} & \textbf{3.57} & \textbf{18.8} \\
         \midrule
        \multirow{2}{*}{VP} & SB & 2.94 & 4.07 & 3.40 & 3.50 & 17.8 \\
        &RSB& \textbf{2.99} & \textbf{4.11} & \textbf{3.46} & \textbf{3.55} & \textbf{18.4} \\
        \bottomrule
    \end{tabular}
\end{table}

\paragraph{Noise Schedule} We further investigate the performance of RSB using VE and VP schedules, and Table~\ref{tab:ablation_noise_schedule} reports the denoising performance under different schedules:
(1) VE: RSB matches PESQ and improves the remaining metrics. Although PESQ remains the same as that of SB, RSB improves signal fidelity, as evidenced by a 1.0 dB increase in SI-SDR, while simultaneously achieving the highest perceptual scores for CSIG, CBAK, and COVL.
(2) VP: RSB exhibits superior performance compared to its SB counterpart, achieving an SI-SDR of 18.4 dB, which is a 0.6 dB improvement over that of SB. The intrusive metrics for perceptual quality are also consistently higher for RSB.

Overall, RSB delivers robust improvements under both VE and VP schedules, with particularly pronounced gains in SI-SDR. Notably, the VE schedule achieves the best overall performance and is hence adopted as the default.

\subsection{Listening Test Results}
\begin{figure}[t]
\centering
\includegraphics[width = 0.9 \columnwidth]{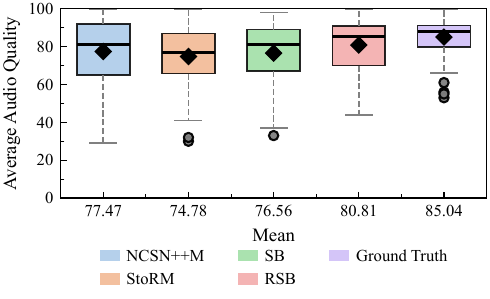}
\vspace{-5mm}
\caption{Results of the MUSHRA listening test. In the boxplots, the mean is denoted by ($\mydiamond$), while the median is marked by ($\boldsymbol{-}$). The box spans the interquartile range, spanning from the 25\textsuperscript{th} to the 75\textsuperscript{th} percentile. The whiskers depict the maximum and minimum values, excluding any outliers. }
\label{fig:mushra_listening_test}
\end{figure}

Beyond the above objective evaluation, subjective evaluation is also essential to assess perceptual quality. We conduct a MUSHRA listening test and present the subjective audio quality ratings for predictive NCSN++M and several generative methods as boxplots in Fig.~\ref{fig:mushra_listening_test}. RSB achieves the highest subjective score, outperforming the baselines NCSN++M, SB, and StoRM. In addition to the improved mean, RSB also exhibits a higher median and a more concentrated interquartile range, indicating more consistent perceptual quality across utterances and listeners. Notably, RSB substantially reduces the subjective-quality gap to the clean reference, indicating that its enhanced outputs are perceived as more similar to clean recordings.

\subsection{Performance in Downstream ASR Task}
\begin{figure}[t]
\centering
\includegraphics[width = 0.7 \columnwidth]{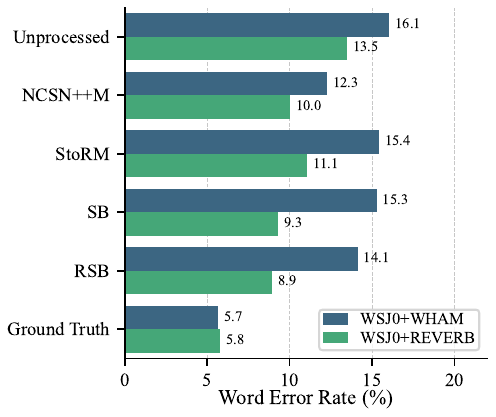}
\vspace{-5mm}
\caption{Word Error Rate (WER) comparison on the downstream ASR task. On each dataset, all utterances are transcribed via Whisper, and the reported WER is the mean across all utterances.}
\vspace{-3mm}
\label{fig:asr_wer}
\end{figure}
As RSB is tailored for high-fidelity SE, we further verify its effectiveness in preserving linguistic content by evaluating downstream ASR performance. Specifically, for WSJ0+WHAM and WSJ0+REVERB, we transcribe the enhanced utterances from all methods using the same Whisper model and report the WER. As illustrated in Fig.~\ref{fig:asr_wer}, on WSJ0+WHAM, NCSN++M achieves the lowest WER of $12.3\%$, which is consistent with the strong denoising performance of the distortion-optimal estimate as reflected by the signal-fidelity metrics in Table~\ref{tab:wsj_main_results}. RSB obtains the best WER among diffusion-based generative models at $14.1\%$, suggesting that DP Perturbation helps preserve the fidelity of the final generated speech. On WSJ0+REVERB, RSB achieves the lowest WER of $8.9\%$. This result is consistent with reduced temporal smearing and residual room effects, while the qualitative visualization in Fig.~\ref{fig:spectrogram_visualization} further indicates that RSB restores sharp harmonic structures and clear spectral envelopes. The results on the two datasets indicate the effectiveness of RSB in preserving linguistic content while suppressing degradation.

\subsection{Evaluating the Distortion-Perception Tradeoff}
The DP tradeoff, also referred to as the fidelity-realism tradeoff, is a fundamental challenge in SE. We evaluate its empirical manifestation using PESQ as an intrusive measure of perceptual quality and SI-SDR as a measure of signal fidelity. We conduct two complementary experiments to investigate the tradeoff. First, we vary the number of sampling steps and report the resulting DP performance.
Second, we compare several strategies designed to improve the DP tradeoff by injecting the predictive estimate into the SB framework, including our proposed DP Perturbation.

\begin{figure}[!t]
\centering
\includegraphics[width=0.9 \columnwidth]{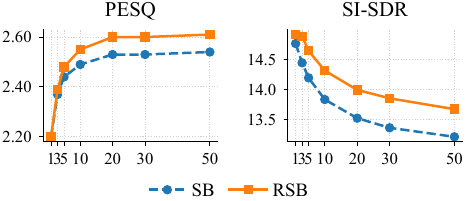}
\vspace{-3mm}
\caption{Denoising performance on the WSJ0+WHAM dataset as a function of the number of sampling steps.}
\vspace{-3mm}
\label{fig:sampling_step_metrics}
\end{figure}

\paragraph{Number of Sampling Steps}
In Fig.~\ref{fig:sampling_step_metrics}, we report the speech denoising performance of SB and RSB under different numbers of sampling steps on the WSJ0+WHAM dataset. PESQ rises sharply in early steps and quickly saturates, indicating fast convergence in perceptual quality, while SI-SDR gradually decreases as perceptual quality increases, reflecting the tradeoff between fidelity and realism. Notably, RSB consistently achieves higher PESQ and better preserves SI-SDR across all steps, with a markedly slower degradation. This demonstrates that while perceptual quality improves at the expense of reconstruction accuracy, RSB attains high perceptual quality earlier while maintaining distortion performance, thereby achieving a more favorable DP tradeoff than SB.

\begin{table}[!t]
    \centering
    \caption{Denoising Performance on the VoiceBank+DEMAND dataset using different DP tradeoff strategies. }
    \label{tab:dp_tradeoff_strategies}
    \footnotesize
    \begin{tabular}{@{}l|cccc|c@{}}
        \toprule
        Method & PESQ$\uparrow$  & CSIG$\uparrow$  & CBAK$\uparrow$ & COVL$\uparrow$  & SI-SDR$\uparrow$  \\
        \midrule
        M1 & \textbf{3.04} & 4.06 & 3.45 & 3.56 & 17.8 \\
        M2 & 3.02 & 4.05 & 3.45 & 3.54 & 18.0 \\
        M3 & 2.96 & 4.03 & 3.40 & 3.50 & 17.7 \\
        M4 & 2.95 & 4.01 & 3.40 & 3.48 & 17.8 \\
        M5 & \textbf{3.04} & \textbf{4.09} & \textbf{3.50} & \textbf{3.57} & \textbf{18.8} \\
        \bottomrule
    \end{tabular}
\end{table}

\paragraph{Tradeoff Strategies}
Table~\ref{tab:dp_tradeoff_strategies} compares different strategies for improving the DP tradeoff on the VoiceBank+DEMAND dataset. The vanilla SB (M1) achieves strong perceptual quality and serves as the baseline for other variants (M2–M5). M2 uses the predictive estimate as the starting point of SB sampling, improving distortion performance. M3 utilizes the predictive estimate as the conditioning of SB, whereas M4 considers it both as the starting point and conditioning, adopting a strategy similar to that of StoRM~\cite{lemercier2023storm}. The results show that M3 and M4 underperform M2 in both distortion and perception quality. This indicates that the predictive estimate may constrain SB too strongly and limit exploration of perceptually favorable solutions, leading to suboptimal tradeoffs. In contrast, our DP Perturbation (M5) achieves the best overall DP performance, improving speech fidelity while maintaining perceptual quality.

It is worth noting that, while M2--M4 require the predictive model to generate the predictive estimates on the fly during inference, RSB only relies on the predictive model during training and eliminates this dependency at inference time, offering a simple yet effective framework.

\subsection{Effectiveness in Mitigating Exposure Bias}
To better understand exposure bias and compare SB and RSB in mitigating this bias, we analyze the evolution of prediction errors along the sampling trajectory on the WSJ0+WHAM dataset. As illustrated in Fig.~\ref{fig:pred_error_trajectory}~(a), we plot the per-step prediction error over 50 sampling steps. The error curve of vanilla SB grows at an accelerating rate as sampling progresses, underscoring the need to mitigate exposure bias. In contrast, RSB yields a markedly flatter curve in early steps, indicating slower error accumulation. The rise in prediction error at later sampling steps is also consistent with our observation that the model increasingly prioritizes distributional realism over reconstruction fidelity. Consequently, RSB achieves lower prediction errors, delivering lower distortion and higher fidelity. Moreover, in Fig.~\ref{fig:pred_error_trajectory}~(b), RSB achieves better PESQ and SI-SDR scores than SB. This confirms that RSB not only mitigates exposure bias effectively but also produces higher-quality solutions.

\subsection{\texorpdfstring{Sensitivity to the Choice of $D_\phi$}{Sensitivity to the Choice of D-phi}}
\begin{table}[!t]
    
    \centering
    \caption{Denoising performance on the VoiceBank+DEMAND dataset with different choices of $D_\phi$ for constructing the distortion-optimal reference $X^*$. Except for the ``--'' row, each row reports the final RSB performance rather than the standalone performance of $D_\phi$.}
    \label{tab:ablation_dphi_reference}
    \footnotesize
    \begin{tabular}{@{}l|cccc|c@{}}
        \toprule
        $D_\phi$ & PESQ$\uparrow$ & CSIG$\uparrow$ & CBAK$\uparrow$ & COVL$\uparrow$ & SI-SDR$\uparrow$ \\
        \midrule
        - & 3.04 & 4.06 & 3.45 & 3.56 & 17.8 \\
        NCSN++M & 3.04 & 4.09 & 3.50 & 3.57 & \textbf{18.8} \\
        MetricGAN+ & 3.02 & 4.04 & 3.46 & 3.53 & 18.2 \\
        SEMamba & \textbf{3.08} & 4.10 & 3.52 & 3.60 & 18.4 \\
        MP-SENet & \textbf{3.08} & \textbf{4.12} & \textbf{3.53} & \textbf{3.61} & \textbf{18.8} \\
        \bottomrule
    \end{tabular}
    \vspace{0.5ex}
    
\end{table}

To isolate sensitivity to the choice of $D_\phi$, we keep the backbone architecture and training procedure fixed, and vary only the predictive model used for offline generation of the distortion-optimal references $X^*$. The row without $D_\phi$ reports vanilla SB and therefore measures the contribution of introducing a predictive reference. This controlled design evaluates whether $D_\phi$ supplies useful fidelity-preserving guidance for DP Perturbation.

Table~\ref{tab:ablation_dphi_reference} shows that a stronger $D_\phi$ can benefit the final RSB performance, although the resulting gain may be limited. NCSN++M raises SI-SDR from 17.8 to 18.8~dB while preserving PESQ at 3.04 and improving the composite quality scores. SEMamba and MP-SENet increase PESQ to 3.08. MP-SENet further improves CSIG, CBAK, and COVL while matching the 18.8~dB SI-SDR of NCSN++M, whereas SEMamba reaches 18.4~dB. MetricGAN+ also improves SI-SDR over vanilla SB, from 17.8 to 18.2~dB, but yields a weaker overall tradeoff than the other reference choices.

These results support the intended role of $D_\phi$ in producing the distortion-optimal reference $X^*$ that provides fidelity-preserving guidance during RSB training. A stronger $D_\phi$ can produce a more reliable reference and thereby improve the final RSB performance. Once the reference is sufficiently reliable, however, the extent to which further improvements in $D_\phi$ translate into better RSB performance is constrained by the modeling capability of the SB model, whose learned SB dynamics ultimately determine the quality of enhanced speech.

\section{Conclusion}\label{sec:conclusion}
In this paper, we identify two practical limitations of SB: the fidelity-realism tradeoff and exposure bias. To address them, we propose the Regularized Schr\"odinger Bridge (RSB), a generative framework for high-fidelity SE that regularizes SB training with our DP Perturbation. Extensive experiments demonstrate that RSB improves speech fidelity while maintaining strong perceptual quality. The listening test and ASR evaluation validate its practical performance. Furthermore, RSB is readily applicable to a wide range of inverse problems beyond SE, such as image denoising and deblurring. Finally, we acknowledge that a limitation of this work is the lack of an in-depth theoretical analysis of the form of prediction drift. Such an analysis could further guide the design of perturbation strategies and the choice of interpolation schedules. We consider this a meaningful direction for future work.

\balance
\bibliographystyle{IEEEtran}
\bibliography{reference}

\end{document}